\pdfoutput=1
\documentclass[11pt]{article}

\usepackage[]{ACL2023}

\usepackage{times}
\usepackage{latexsym}

\usepackage[T1]{fontenc}
\usepackage[utf8]{inputenc}

\usepackage{microtype}

\usepackage{inconsolata}

\usepackage{microtype}
\usepackage{booktabs}
\usepackage{tablefootnote}
\usepackage{xcolor}
\usepackage{multirow}
\usepackage{amssymb}
\usepackage{pifont}

\usepackage{graphicx}
\usepackage{hyperref}
\usepackage{xspace}
\usepackage{caption}
\usepackage{subcaption}
\usepackage{array}
\newcolumntype{L}[1]{>{\raggedright\let\newline\\\arraybackslash\hspace{0pt}}m{#1}}
\newcolumntype{C}[1]{>{\centering\let\newline\\\arraybackslash\hspace{0pt}}m{#1}}
\newcolumntype{R}[1]{>{\raggedleft\let\newline\\\arraybackslash\hspace{0pt}}m{#1}}

\def\corpus/{IndicCorp}
\def\benchmark/{IndicXTREME}
\def\model/{IndicBERT}

\newcommand{\cmark}{\ding{51}}%
\newcommand{\xmark}{\ding{55}}%

\title{Towards Leaving No Indic Language Behind: Building Monolingual Corpora, Benchmark and Models for Indic Languages}

\author{
        Sumanth Doddapaneni$^{1,2}$\thanks{~~Corresponding authors: Sumanth Doddapaneni (\href{mailto:dsumanth17@gmail.com}{dsumanth17@gmail.com}), Mitesh M. Khapra (\href{mailto:miteshk@cse.iitm.ac.in}{miteshk@cse.iitm.ac.in})}  \hspace{0.2cm} Rahul Aralikatte$^{4,5}$ \\ 
        \textbf{Gowtham Ramesh}$^{2}$  \hspace{0.2cm} \textbf{Shreya Goyal}$^{2}$ \\ \textbf{Mitesh M. Khapra}$^{1,2}$ \hspace{0.2cm} \textbf{Anoop Kunchukuttan}$^{1,2,3}$ \hspace{0.2cm}
         \textbf{Pratyush Kumar}$^{1,2,3}$
    \\
    $^1$Indian Institute of Technology, Madras \hspace{0.2cm} 
    $^2$AI4Bharat \\ 
    $^3$Microsoft \hspace{0.2cm} 
    $^4$Mila - Quebec AI Institute \hspace{0.2cm}
    $^5$McGill University \hspace{0.2cm} 
}

\begin{document}
\maketitle
\begin{abstract}

Building Natural Language Understanding (NLU) capabilities for Indic languages, which have a collective speaker base of more than one billion speakers is absolutely crucial. In this work, we aim to improve the NLU capabilities of Indic languages by making contributions along 3 important axes (i) monolingual corpora (ii) NLU testsets  (iii) multilingual LLMs focusing on Indic languages. Specifically, we curate the largest monolingual corpora, IndicCorp, with 20.9B tokens covering 24 languages from 4 language families - a 2.3x increase over prior work, while  supporting 12 additional languages. Next, we create a human-supervised benchmark, IndicXTREME, consisting of nine diverse NLU tasks covering 20 languages. Across languages and tasks, IndicXTREME contains a total of 105 evaluation sets, of which 52 are new contributions to the literature. To the best of our knowledge, this is the first effort towards creating a standard benchmark for Indic languages that aims to test the multilingual zero-shot capabilities of pretrained language models. Finally, we train IndicBERT v2, a state-of-the-art model supporting all the languages. Averaged across languages and tasks, the model achieves an absolute improvement of 2 points over a strong baseline. The data and models are available at \url{https://github.com/AI4Bharat/IndicBERT}.

\end{abstract}

\section{Introduction}

Recent advances in Natural Language Understanding are largely driven by pretrained multilingual models \cite{conneau-etal-2020-unsupervised, xue-etal-2021-mt5, DBLP:journals/corr/abs-2107-00676}. One of the advantages of such models is that they can potentially reduce the performance gap between high and low-resource languages through zero-shot knowledge transfer \cite{pmlr-v119-hu20b, liang-etal-2020-xglue}. However, in practice, the benefits of such models are still skewed towards high-resource languages due to 3 main reason as outlined below.

First, current multilingual models often have a poor representation of low-resource languages. For example, out of the 22 languages listed in the  8$^{th}$ schedule of the Indian constitution, only 15 languages are supported by the popular XLM-R model \cite{conneau-etal-2020-unsupervised}. This is mainly due to the non-availability of pretraining data for languages like Bodo, Dogri, Kashmiri, etc. in large multilingual corpora such as CC-100 \citep{conneau-etal-2020-unsupervised}, or mC4 \citep{xue-etal-2021-mt5}. Hence, dedicated efforts towards collecting pretraining data for these languages by discovering and crawling language-specific sources are needed.

\begin{table}[]
\centering
\tiny{
\setlength{\tabcolsep}{6pt}
\begin{tabular}{R{1.7cm}>{\centering}m{0.7cm}>{\centering}m{1.2cm}>{\centering}m{0.5cm}C{1.2cm}}
\toprule
  & \bf XTREME & \bf XTREME-R & \bf XGLUE & \bf IndicXTREME \\
 \midrule
\#Indic lang. tasks & 25 & 28 & 5 & 105 \\
Avg. \#test ins./task & 1691.9 & 1842.7 & 3845.6 & 2008 \\
\midrule
 & \bf Wikipedia & \bf CC-100 & \bf mC4 & \bf IndicCorp \\
 \midrule
\#Indic lang. & 20 & 15 & 12 & 23 \\
\#Indic lang. tokens & 0.2B & 5.0B & 20.2B\tablefootnote{Note that while the number of tokens in mC4 is larger than that in IndicCorp, recent studies \cite{kreutzer-etal-2022-quality} have shown that mC4 contains a significant amount of offensive and pornographic content. Further, it is often the case that the content does not belong to the designated language. This is mainly because the data is not crawled from verified URLs. In contrast, in IndicCorp we make a conscious choice to crawl content only from human-verified URLs.} & 14.4B \\
Verified source URLs & \checkmark & \xmark & \xmark & \checkmark~ \\
\midrule
 & \bf mBERT & \bf XLM-R & \bf MuRIL & \bf IndicBERT \\
 \midrule
 \#Indic / \#Total langs. & 11/104 & 15/110 & 16/17 & 23/24 \\
 Fertility ($\downarrow$) & 2.8 & 2.2 & 1.7 & 1.7 \\
\bottomrule
\end{tabular}
}
\caption{A comparison of existing benchmarks, pretraining corpora, and multilingual language models with \benchmark/, \corpus/, and \model/ respectively, in the context of Indic languages. In row 2, the average is computed only for Indic languages.}
\label{tab:gist}
\end{table}

Second, even for low-resource languages supported by existing multilingual models, the size of pretraining data is much smaller than that of English and other resource-rich languages \cite{xue-etal-2021-mt5}. Due to this disparity, low-resource languages get a very poor share of the model's capacity and vocabulary, and thus the performance on these languages is poor \cite{conneau-etal-2020-unsupervised}. Indeed, a few recent efforts \cite{kakwani-etal-2020-indicnlpsuite, DBLP:journals/corr/abs-2103-10730, dabre-etal-2022-indicbart, reid-etal-2021-afromt} show that multilingual models trained using pretraining data from a smaller set of related languages leads to better performance on downstream tasks than large scale models which support many languages. Hence, there is a need for training language models only on Indic languages thereby ensuring that the model capacity is not dominated by unrelated high-resource languages. 

The third reason is the poor representation of these languages in existing evaluation benchmarks. For example, in the XTREME-R \cite{DBLP:conf/emnlp/RuderCBSFF00GNJ21} benchmark, out of the 10 tasks only three contain evaluation data for more than two Indic languages. Further, the maximum number of Indic languages for any task is just seven. In effect, 15 of the 22 constitutionally recognized Indic languages have no representation in XTREME-R for any task. Thus, a human supervised evaluation benchmark tailored for Indic, and other low-resource language families is essential for furthering inclusivity and equity in NLP research \cite{DBLP:journals/corr/abs-2205-12676}. 

In this work, we make contributions toward addressing all the three challenges. We focus on the 22 languages listed in the 8$^{th}$ schedule of the Indian constitution spanning 4 language families and spoken by over a billion speakers (8 of these languages being amongst the top-20 most spoken languages globally). Some of these languages are also widely spoken and/or are official languages in neighbouring countries \textit{viz.,} Bangladesh, Nepal and Pakistan. Our first contribution towards serving these languages is to release \corpus/ v2, the largest collection of corpora for languages spanning 4 Indic language families with 20.9 Billion tokens and 1.1 Billion sentences. Table \ref{tab:gist} shows a comparison of \corpus/ v2 with existing collections of monolingual corpora. As is clear, \corpus/ not only supports more Indic languages but also improves upon the data for languages supported in existing collections (e.g., $\times$2.3 improvement over IndicCorp v1 with 12B new tokens). Our second contribution is \model/ v2, a multilingual LM pretrained on \corpus/ v2 and supporting the largest number of Indic languages compared to existing models such as XLM-R, MuRIL, and IndicBERT v1.

Our third, and perhaps, the most important contribution is \benchmark/, a human supervised benchmark containing evaluation sets for nine diverse tasks with each task covering 7-18 Indic languages per task. These include five classification tasks, two structure prediction tasks, one QA task, and one text retrieval task. Of the total 105 evaluation sets, summed across languages and tasks, 52 have been newly created as a part of this benchmark. All the newly added evaluation sets have been created manually with the help of in-house language experts with several years of experience in language annotation and translation. The datasets for three tasks, \textit{viz.}, NER, QA, and paraphrase detection were created from scratch without any translation from English sources. We consciously make an effort to include languages spanning all the classes from the inclusion taxonomy introduced in \citet{joshi-etal-2020-state}. According to their classification (Table \ref{tab:lang-classes}), nine languages in \benchmark/ are the so-called ``Left-Behinds'', the most ignored, with exceptionally minimal resources. Only three are ``Winners'', the high-resource languages, which have a dominant online presence with industry and government investments.

Using \benchmark/, we evaluate \model/ and show that it outperforms strong baselines on 7/9 evaluation tasks. We also do a series of ablation tests to show that (i) the translation language modeling (TLM) objective slightly improves zero-shot performance when high-quality parallel data is used,  (ii) using noisy parallel data during pretraining leads to sub-optimal zero-shot performance, (iii) using in-language-family development sets allows better model selection, and (iv) zero-shot transfer via Hindi, as opposed to English, leads to better performance. All the datasets, code, and models developed as a part of this work will be open-sourced.
All the datasets and models developed as a part of this work are available at \url{https://ai4bharat.iitm.ac.in/language-understanding}.

\begin{table*}[]
\centering
\small
\begin{tabular}{rllrrcccc}
\toprule
Task Category & Dataset & Task & |Dev| & |Test| & Method & |Lang.| & Metric & Domain \\
\midrule
\multirow{5}{*}{Classification} & IndicSentiment & \begin{tabular}[c]{@{}l@{}}Sent. \\ Classification\end{tabular} & 156 & 1000 & HA & 13 & Acc. & Reviews \\
 & IndicXNLI & NLI & 2490 & 5010 & MT$^\zeta$ & 12 & Acc. & Misc \\
 & IndicCOPA & Reasoning & - & 500 & HA & 18 & Acc. & Misc \\
 & IndicXPara & \begin{tabular}[c]{@{}l@{}}Sent. \\ Equivalance\end{tabular} & - & 2002 & HA & 10 & Acc. & Misc. \\
 & M-Intent & Intent & 2033 & 2974 & HA & 7 & Acc & Spoken \\
 \midrule
\multirow{2}{*}{\begin{tabular}[c]{@{}l@{}}Structure\\ Prediction\end{tabular}} & Naamapadam & NER & 52-13460 & 607-1080 & HA & 9 & F1 & News \\
 & M-SlotFill & Slot Filling & 2033 & 2974 & HA & 7 & F1 & Spoken \\
 \midrule
QA & IndicQA & Span Extraction & - & 1517-2017 & HA & 11 & F1 & Wiki. \\
\midrule
Retrieval & FLORES & Sent. Retrieval & - & 1012 & HA & 18 & Acc. & Wiki++ \\
\midrule
\end{tabular}
\caption{A summary of the tasks in \benchmark/. |Lang| denotes the number of languages for which test sets are available. |Test| is the size of the test sets in each language. |Dev| is the size of in-language development sets, if available. HA, \& MT stand for `Human Annotated' \& `Machine Translation' respectively. The `M' in M-Intent and M-SlotFill refers to the MASSIVE dataset \cite{DBLP:journals/corr/abs-2204-08582}. $\zeta$ - Human verification is in progress, please refer to Appendix \ref{app:xnli-cleaning}}
\label{tab:tasks}
\end{table*}

\section{Related Work}
The ability of multilingual models to do zero-shot transfer is often limited to typological cousins inside language families \cite[Section 2]{ponti-etal-2021-minimax}. This has spurred coordinated research efforts for underrepresented languages, such as Indic languages. Recent works in this domain can be broadly classified into the following three broad areas.

\subsection{Resources}
The data resource used most often for pretraining models in Indic languages is Wikipedia. Though it has high-quality text, Indic Wikis are sparsely populated\footnote{Apart from Hindi, which has 153,000 articles as of November 2022 all others have few thousand articles.}. Corpora derived from CommonCrawl like CC100 \cite{conneau-etal-2020-unsupervised} and mC4 \cite{xue-etal-2021-mt5} are a popular source for major Indian languages. However, this text is often noisy and contains offensive content \cite{DBLP:journals/tacl/KreutzerCWWEUTS22}. IndicCorp v1 \cite{kakwani-etal-2020-indicnlpsuite} is the first effort to curate a pretraining corpus exclusively for Indic languages. In this work, we build upon IndicCorp v1 to include more languages as well as crawl more data for existing languages. 

\subsection{Models}
Most multilingual pretrained language models and their variants like mBERT \cite{DBLP:conf/naacl/DevlinCLT19}, mT5 \cite{xue-etal-2021-mt5}, and XLM \cite{DBLP:conf/nips/ConneauL19} are trained on major Indic languages. However, it is difficult to get optimum performance from these models on Indic tasks as they have to compete for model capacity with other high-resource languages \cite{conneau-etal-2020-unsupervised, DBLP:journals/corr/abs-2205-12676}. Indic family-specific models like MuRIL \cite{DBLP:journals/corr/abs-2103-10730} and IndicBERT v1 \cite{kakwani-etal-2020-indicnlpsuite} do much better on such tasks than the aforementioned models. 

\subsection{Benchmarks}
Benchmarks like GLUE \cite{wang-etal-2018-glue} and SuperGLUE \cite{DBLP:conf/nips/WangPNSMHLB19} have driven research on multitask models for English and IndicGULE \cite{kakwani-etal-2020-indicnlpsuite} has been created to benchmark performance on Indic languages. Similarly, there have been multiple efforts to drive research on crosslingual, multitask models. Important among them are XGLUE \cite{liang-etal-2020-xglue}, XTREME \cite{pmlr-v119-hu20b}, and XTREME-R \cite{DBLP:conf/emnlp/RuderCBSFF00GNJ21}. In order to accommodate a diverse set of languages, these benchmarks have a limited representation of Indic languages. Also, most evaluation sets are automatically translated or generated which is known to have problems \cite{Vanmassenhove2021MachineTE}. In this work, we aim to fill this gap by presenting an Indic family-specific evaluation benchmark consisting of 9 tasks with human-created or human-translated test sets.

\section{\benchmark/~}
The \benchmark/~ benchmark includes 9 tasks that can be broadly grouped into sentence classification (5), structure prediction (2), question answering (1), and sentence retrieval (1). Since the benchmark is designed to evaluate models in a zero-shot setting, we only create test sets. Table \ref{tab:tasks} gives a summary of the testsets in \benchmark/.

\subsection{New Contributions}

\paragraph{IndicCOPA} 
\label{ss:xcopa}
We manually translate the COPA \citep{roemmele2011choice} test set into 18 Indic languages to create IndicCOPA. The premise and the choices from the original dataset are randomized and assigned to translators to avoid any bias. Once translated, the sentences are re-grouped. For fine-tuning, we use the English Social IQA dataset \cite{sap-etal-2019-social}.

\paragraph{IndicQA}
\label{ss:indicqa}
We introduce IndicQA, a manually curated cloze-style reading comprehension dataset that can be used for evaluating question-answering models in 11 Indic languages. The context paragraphs are chosen from Wikipedia articles whose topics are closely related to Indic culture, history, etc. The dataset consists of 18,579 questions out of which 13,283 are answerable. A language-wise breakdown of the numbers can be seen in Table \ref{tab:indicqa_stats} in Appendix \ref{apx:indicqa}. For more details about the collection process and annotation guidelines, see Appendix \ref{app:indicqa_guidelines}. For fine-tuning of baseline models, we use the English SQuAD \cite{rajpurkar-etal-2016-squad} dataset. 

\paragraph{IndicXParaphrase}
\label{ss:paraphrase}
We take 1001 English sentences from \citet{DBLP:journals/corr/abs-2203-05437} with a mean sentence length of 17 words. We auto-translate these sentences into 10 languages using the IndicTrans translation model \citep{DBLP:journals/tacl/RameshDBJASSDJK22}. Human annotators then verify (and correct, if required) these translations. Next, the annotators manually create paraphrases and non-paraphrases for each translated sentence. This results in 1001-way parallel \textit{<sentence, paraphrase, non-paraphrase>} triplet in each of the 10 languages, where the {\it sentences} are shared across languages. The annotators are provided with strict guidelines to ensure the quality of the (non-)paraphrases. See Appendix \ref{app:indic_paraphrase} for more details about the annotation process. Contrary to prior works like \citet{yang-etal-2019-paws}, we do not use back-translation or other noisy alignment methods to create non-paraphrases. For fine-tuning, we use the English part of the PAWS-X \cite{yang-etal-2019-paws}.

\paragraph{IndicSentiment}
\label{ss:sentiment}
In general, product reviews are one-dimensional and a vast majority of the reviews are highly polarized which makes classification easy. This results in models performing poorly on nuanced reviews. Therefore in this dataset, we ask annotators to create synthetic reviews for real products. We curate a list of aspects for each product category and ask the annotators to write reviews that talk about a subset of those aspects. All the reviews are first written in English and then manually translated to 13 Indic languages, thus making it a 13-way parallel dataset. More information about annotation guidelines can be found in Appendix \ref{app:sentiment}. For fine-tuning, we use the English Amazon Multilingual Reviews dataset \citep{DBLP:conf/emnlp/KeungLSS20}.

\subsection{Other Datasets}

\paragraph{IndicXNLI}
\label{ss:indicxnli}
This dataset, already proposed in \cite{DBLP:journals/corr/abs-2204-08776} released an automatically translated version of XNLI \cite{DBLP:conf/emnlp/ConneauRLWBSS18} in 11 Indic languages. Though the translations are generally good, there are certain quality issues that are a result of the dataset containing text that is a transcription of spoken language. This results in the translations being structurally and semantically incorrect. In this work, we manually verify the translations of some parts of the test set and make changes where necessary. Due to cost and time constraints, we could not verify the entire test set. Please see Table \ref{app-tab:xnli-cleaning} in Appendix \ref{app:xnli-cleaning} to see the number of instances that were manually verified and corrected across languages. We plan to continue this effort and correct/verify the entire test set over a period of six months. For fine-tuning, we use the MultiNLI dataset \citep{williams-etal-2018-broad}.

\paragraph{Naamapadam}
\label{ss:naamapadam}
This NER dataset was proposed in \citet{mhaske2022indicner}\footnote{\url{https://huggingface.co/datasets/ai4bharat/naamapadam}} with manually curated testsets for nine Indic languages. The testsets have been created using the following process: (i) for an English-Indic language parallel sentence pair, the English sentence was NER tagged using an off-the-shelf model, (ii) the NER tags were automatically projected to the Indic language sentence via word alignments, and (iii) the tags in the Indic sentence were verified and corrected by annotators. The annotations follow the standard IOB2 format. For training and validation, we use the CoNLL-2003 dataset \cite{tjong-kim-sang-de-meulder-2003-introduction}.

\paragraph{FLORES}
\label{ss:flores}
To evaluate the retrieval capabilities of models, we include the Indic parts of the FLORES-101/200 dataset \citep{DBLP:journals/tacl/GoyalGCCWJKRGF22,DBLP:journals/corr/abs-2207-04672} to \benchmark/. This is an n-way parallel dataset containing 1012 sentences manually translated into 18 Indic languages. We do not perform any fine-tuning and use mean-pooled representations from the final layer of the models as sentence embeddings.

\paragraph{MASSIVE}
\label{ss:massive}
This intent classification and slot-filling dataset proposed by \citet{DBLP:journals/corr/abs-2204-08582} is created using user queries collected by Amazon Alexa. The dataset contains 60 intents and 55 slot types and is available in 51 languages. We take a subset of it consisting of seven Indic languages to be part of \benchmark/. We use the English train and validation sets for training baseline models.

We reemphasise that \textbf{ALL} the evaluation sets included in IndicXTREME were created with human supervision. In other words, 
they were either translated or post-edited or created or verified by humans.

\section{\corpus/ v2}
In this section, we describe the process followed to build \corpus/ v2, the largest collection of texts for Indic languages consisting of 20.9 billion tokens of which 14.4B tokens correspond to 23 Indic languages and 6.5B tokens of Indian English content curated from Indian websites. Table \ref{tab:corpus_stats} shows the size of the de-duplicated corpus across languages. The current corpus (24 languages) is 2.3$\times$ compared to IndicCorp v1 (12 languages) with the largest increase in Hindi (3.3$\times$). The corpus contains 1.08 billion tokens from the bottom 11 low-resource languages.

\subsection{Data}
With the goal of creating a clean and diverse corpus, we choose news articles as our primary sources. In addition to the sources already discovered by \citet{kakwani-etal-2020-indicnlpsuite}, we identify new sources for more languages through news repositories and automatic web searches. In particular, we determine the most frequent words that occur in a language and use these as queries for automated web searches. We identify URLs of sources that potentially contain content in those languages from the retrieved results. An analysis of the retrieved URLs shows that some of them are noisy with offensive content or machine-generated content. We, therefore, add a filtering stage wherein we ask human annotators to manually verify the URLs. Specifically, each annotator is asked to visit the URL and verify that it is a genuine website containing clean data in the language of interest. Across languages, we find that 1-33\% of the URLs are noisy and we discard them. We then used the open-source toolkit \textit{webcorpus}\footnote{\url{https://gitlab.com/AI4Bharat/NLP/webcorpus}} to crawl the shortlisted URLs. 

\subsection{Post-processing}
\label{ref:corp-postproc}

We process the crawled dumps to produce clean text. We see that the crawls often contain data from other languages. In order to remove such undesired text, we perform language detection-based (LID) filtering at paragraph level using cld3\footnote{\url{https://github.com/google/cld3}} and langdetect\footnote{\url{https://github.com/shuyo/language-detection}} and discard text that is not in the language  of interest. Note that low-resource languages like \texttt{bd} and \texttt{dg} are not supported by the libraries and hence we do not perform LID-based filtering for these languages. 

Previous works suggest that data crawled from the web often contains offensive text \cite{DBLP:journals/tacl/KreutzerCWWEUTS22}. To remove such text from our corpus, we create a list of offensive words and phrases in 17 languages with the help of in-house annotators. In a parallel approach, a similar list of offensive words was released for 209 languages by \citet{DBLP:journals/corr/abs-2207-04672}. We merge these two lists to create a comprehensive blacklist of words for all languages in the corpus. This list is used to filter text containing offensive content reducing the corpus size from 23.1 billion to 20.9 billion tokens. Following \citet{kakwani-etal-2020-indicnlpsuite}, we add data from Wikipedia and OSCAR \cite{OrtizSuarez2019AsynchronousPF} to our final corpus. 

\begin{table}
  \small
  \centering
\begin{tabular}{rcc|rcc}
\toprule
L & v1 & v2 & L & v1 & v2 \\
\midrule
as & 32.6 & 67 & ml & 721 & 931  \\
brx & - & 2.5 & mni & - & 0.6  \\
bn & 836 & 926 & mr & 551 & 795  \\
doi & - & 0.1 & ne & - & 852 \\
en & 1220 & 6501 & or & 107 & 122 \\
gom & - & 31.9 & pa & 773 & 732 \\
gu & 719 & 901 & sa & - & 125 \\
hi & 1860 & 6107 & sat & - & 4 \\
kha & - & 46 & sd & - & 13.2 \\
kn & 713 & 875& ta & 582 & 476 \\
ks & - & 0.06 & te & 674 & 731 \\
mai & - & 13.7 & ur & - & 667 \\
\midrule
\multicolumn{1}{l}{} & \multicolumn{1}{l}{} & \multicolumn{1}{l}{} & \multicolumn{1}{l}{Total} & \multicolumn{1}{l}{8789} & \multicolumn{1}{l}{20920} \\
\bottomrule
\end{tabular}
  \caption{Comparison of the number of tokens (in Millions) in each language of IndicCorp v1 vs. v2.}
  % All numbers are in millions.}
  \label{tab:corpus_stats}
\end{table}

\section{IndicBERT v2}
This section describes the various aspects of training \model/, a language model trained on \corpus/ and evaluated on \benchmark/. In our experiments, we train with BERT architecture and ablate on objective functions and training data. Compared  to IndicBERT v1 \cite{kakwani-etal-2020-indicnlpsuite}, trained on the smaller ALBERT \citep{DBLP:conf/iclr/LanCGGSS20} architecture, this version has $\sim$7.5x  more parameters and is able to transfer across languages in zero-shot settings. The model has 278M parameters and supports all 24 languages in \corpus/. 

\paragraph{Training Objectives}
We experiment with two objective functions: Masked Language Modeling \citep[][MLM]{DBLP:conf/naacl/DevlinCLT19} and Translation Language Modeling \citep[][TLM]{DBLP:conf/nips/ConneauL19}. We use the document-level data created as part of \corpus/ for MLM objective training. Pretraining hyperparameters are listed in Appendix \ref{app:pret-hparams}.

\paragraph{Data}
As mentioned in Section \ref{ref:corp-postproc}, we merge data from \corpus/ v2 with Indic language data from Wikipedia and OSCAR. For MLM, we use these monolingual corpora spanning 24 languages, 5 language families, and 13 scripts. 
For TLM, we use language-parallel data from two sources: mined data from Samanantar corpus \citep{DBLP:journals/tacl/RameshDBJASSDJK22}, and machine-generated English translations of the entire \corpus/. We use IndicTrans \citep{DBLP:journals/tacl/RameshDBJASSDJK22} for all translations. We are limited in our ability to generate parallel sentences since IndicTrans supports only 11 of the 24 languages in \corpus/. We perform ablations by training models on various subsets of this data as discussed in Section \ref{sec:model-ablations}. Since data distribution across languages is skewed (Fig. \ref{fig:upsampling} in Appendix \ref{apx:data-dist}), we follow \citet{DBLP:journals/corr/abs-2103-10730} to upsample the underrepresented languages with 0.3 temperature coefficient.

\paragraph{Vocabulary}
We learn a WordPiece \cite{DBLP:journals/corr/WuSCLNMKCGMKSJL16} vocabulary from a uniformly sampled fraction of the upsampled data. We also add special \texttt{<lang-id>} tokens to the vocabulary since \citet{DBLP:journals/tacl/RameshDBJASSDJK22} have shown that training multilingual models with language tokens improve performance. These tokens are prepended to input documents during pretraining. Given that our model supports 24 languages and 13 scripts, we use a vocabulary size to 250K tokens. See Appendix \ref{app:tokenizers} for more details.

\section{Experiments}
We compare IndicBERT v2 with the following LMs - IndicBERT v1 \cite{kakwani-etal-2020-indicnlpsuite}, mBERT \cite{DBLP:conf/naacl/DevlinCLT19}, XLMR \cite{conneau-etal-2020-unsupervised} and MuRIL \cite{DBLP:journals/corr/abs-2103-10730}. We describe our choice of baseline models, and their similarities and differences in Appendix \ref{app:baseline-models}. We then briefly introduce our fine-tuning details and the various ablation studies conducted. 

\subsection{Fine-Tuning}
The pre-trained LM is independently fine-tuned for each task in \benchmark/. We perform zero-shot evaluation by fine-tuning the model on English and testing on the available Indic test sets. The best configuration of the model is chosen based on its performance on the English development set. While most works in literature \cite{DBLP:journals/corr/abs-2103-10730, conneau-etal-2020-unsupervised} use the same hyperparameters for fine-tuning models on various tasks, we find that task-specific hyperparameter-tuning improves performance. For a fair comparsion, we perform hyperparamter-tuning for all the models that we compare with. Our choice of hyperparameters for each task  can be found in Tables \ref{tab:hparams-copa-para}, and \ref{tab:in-lang-hparams} in the Appendix \ref{app:ft-hparams}. Models  are fine-tuned for every task except for the retrieval task, where we directly use the mean pooled sentence representation from the last layer of the pretrained models. 

\subsection{\model/ v2 Ablations}
\label{sec:model-ablations}
We train four flavors of \model/ v2 to understand the role of parallel data and its quality in improving crosslingual performance. The first model is a vanilla BERT style model trained on \corpus/ v2 with the MLM objective. In the other two ablations, we include TLM as an additional objective with different sets  of parallel data. In one ablation, we include parallel data from the Samanantar dataset.\footnote{Samanantar data is sentence-level parallel and is not ideal. But document-level parallel data for Indic languages are scarce.} This corpus contains high-quality translations mined from various sources and supports 11 Indic languages. These models are denoted by (\textbf{+Samanantar}) in the results. Third, we translate the whole \corpus/ v2 to English using IndicTrans and use it as additional parallel data (\textbf{+Back-Trans} in results). Empirically, the quality of these translated parallel data is lower than those of Samanantar, especially for very low-resource languages like Assamese. Finally, to encourage better lexical sharing among languages we convert the scripts from Indic languages to Devanagari (\textbf{IndicBERT-SS}). All Indian languages are derived from the Brahmi script and there exists a 1-1 mapping between characters across different scripts. We convert all the supported languages to Devanagari script using IndicNLP Library \cite{kunchukuttan2020indicnlp}.

\section{Results}
\label{sec:results}

\begin{table*}[h!]
\centering
\small
\begin{tabular}{lccccccccc}
\toprule
 & \multicolumn{5}{c}{Classification} & \multicolumn{2}{c}{Structure Prediction} & \multicolumn{1}{c}{QA} & \multicolumn{1}{c}{Retreival} \\
\cmidrule(lr){2-6} \cmidrule(lr){7-8} \cmidrule(lr){9-9} \cmidrule(lr){10-10}
\multirow{ 2}{*}{Models} & Indic     & Indic & Indic & Indic & MASSIVE & Naama-   & MASSIVE    & Indic & \multirow{ 2}{*}{FLORES} \\
& Sentiment & XNLI  & COPA  & XPara. & (Intent)  & Padam  & (Slotfill) & QA    &  \\
\midrule
IndicBERT v1 & 61.8 & 42.8 & 51.0 & 47.5 & - & 25.3 & - & 10.1 & 1.1 \\
mBERT & 69.5 & 54.7 & 51.7 & 55.2 & 13.2 & 63.0 & 6.2 & 32.9 & 32.3 \\
XLMR & 84.0 & 69.7 & 60.1 & 56.7 & 66.6 & 71.7 & 50.0 & 44.8 & 3.1 \\
MuRIL & 85.1 & 72.4 & 58.9 & \textbf{60.8} & 77.2 & \textbf{74.3} & 57.0 & 48.3 & 52.3 \\
v1-data & 85.7 & 66.4 & 52.4 & 49.6 & 25.8 & 58.3 & 34.4 & 37.6 & 54.9 \\
\midrule
IndicBERT v2 & \textbf{88.3} & 73.0 & 62.7 & 56.9 & 78.8 & 73.2 & 56.7 & 47.7 & 69.4 \\
$\quad$ +Samanantar & \textbf{88.3} & 74.3 & \textbf{63.0} & 57.0 & 78.8 & 72.4 & \textbf{57.3} & 49.2 & 64.7 \\
$\quad\quad$ +Back-Trans. & 87.5 & 69.7 & 53.8 & 50.7 & 77.4 & 71.9 & 54.6 & 42.2 & 68.6 \\
\midrule
IndicBERT-SS & 88.1 & \textbf{73.9} & 64.2 & 56.4 & \textbf{80.7} & 66.6 & \textbf{57.3} & \textbf{49.7} & \textbf{71.2} \\
\bottomrule
\end{tabular}
\caption{Results averaged across \textbf{languages} from the \benchmark/ benchmark. We report F1 scores for Structure Prediction \& QA, and accuracy for the other tasks.}
\label{tab:results}
\end{table*}

\begin{table*}[h!]
\centering
\small
\begin{tabular}{lcccccccc}
\toprule
\multirow{ 2}{*}{Models} & \multicolumn{2}{c}{IndicSentiment} & \multicolumn{2}{c}{Naamapadam} & \multicolumn{2}{c}{MASSIVE (Intent)} & \multicolumn{2}{c}{IndicXNLI} \\
\cmidrule(lr){2-3} \cmidrule(lr){4-5} \cmidrule(lr){6-7} \cmidrule(lr){8-9} 
& in-lg. & in-fam. & in-lg. & in-fam. & in-lg. & in-fam. & in-lg. & in-fam. \\
\midrule
mBERT & 72.9$_{+3.4}$ & 72.9$_{+3.4}$ & 65.8$_{+2.8}$ & 65.2$_{+2.3}$ & 15.1$_{+1.9}$ & 14.7$_{+1.5}$ & 58.4$_{+3.7}$ & 58.4$_{+3.7}$ \\
XLMR & 86.1$_{+2.1}$ & 84.6$_{+0.6}$ & 73.0$_{+1.3}$ & 73.0$_{+1.3}$ & 67.6$_{+1.0}$ & 67.6$_{+1.0}$ & 70.4$_{+0.7}$ & 70.1$_{+0.4}$ \\
MuRIL & 89.3$_{+4.2}$ & 89.2$_{+4.1}$ & 74.3$_{+0.0}$ & 74.1$_{-0.2}$ & 77.3$_{+0.1}$ & 77.5$_{+0.3}$ & 74.0$_{+1.6}$ & 74.0$_{+1.6}$ \\
\midrule
IndicBERT & 92.5$_{+4.2}$ & 92.5$_{+4.2}$ & 73.2$_{+0.0}$ & 73.2$_{+0.0}$ & 79.1$_{+0.3}$ & 79.1$_{+0.3}$ & 73.0$_{+0.0}$ & 72.6$_{+0.4}$ \\
$\quad$ +Samanantar & 92.4$_{+4.1}$ & 92.4$_{+4.1}$ & 72.9$_{+0.5}$ & 72.9$_{+0.5}$ & 79.2$_{+0.4}$ & 78.9$_{+0.1}$ & 74.3$_{+0.0}$ & 74.3$_{+0.0}$ \\
$\quad\quad$ +Back-Trans. & 93.1$_{+5.6}$ & 92.8$_{+5.3}$ & 72.2$_{+0.4}$ & 72.2$_{+0.4}$ & 77.5$_{+0.1}$ & 77.4$_{+0.0}$ & 71.5$_{+0.8}$ & 71.5$_{+0.8}$ \\
\bottomrule
\end{tabular}
\caption{Performance improvement when we use in-language (\textbf{in-lg.}) and in-family (\textbf{in-fam.}) development sets. The results are in the form X$_Y$ where X is the absolute performance metric value, and Y is the performance increase over a model fine-tuned with an English development set. We run this experiment only on those datasets for which an in-family development set is available.}
\label{tab:dev-set-analysis}
\end{table*}

The results for each task in \benchmark/ averaged across languages are shown in Table \ref{tab:results}.

\paragraph{Massively Multilingual vs Indic Models} It is clear that there is no single best model on the benchmark. However, \model/ v2 family of models beat the baselines in 7/9 tasks. The language-specific results for all experiments can be found in Appendix \ref{app:lang-specific}. When averaged across tasks (see Table \ref{tab:lang-wise-results}), \model/ v2 performs  the best on 17/20 languages. On average, the \model/ v2 family of models, outperform other models.

The results show that models trained only on Indic languages perform better since languages do not have to compete for model capacity. We see that \model/ v2 trained only on MLM, by itself performs much better than the standard baselines. The only exception to this is that MuRIL outperforms \model/ v2 in the paraphrase detection and NER tasks. We also see that adding the TLM objective with (i) high-quality parallel data increases the model performance across the board, and (ii) machine-translated data hurts performance.

\begin{table*}[tbh!]
\centering
\small
\begin{tabular}{lcccccccc}
\toprule
\multirow{ 2}{*}{Models} & Indic     & Indic & Indic & Indic & MASSIVE & Naama-   & MASSIVE    & Indic \\
& Sentiment & XNLI  & COPA  & XPara. & (Intent)  & Padam  & (Slotfill) & QA \\
\midrule
\begin{tabular}[c]{@{}l@{}}IndicBERT v2\\ $\quad$ +Samanantar\end{tabular} & 88.3 & 74.3 & \textbf{63.0} & 57.0 & 78.8 & 72.4 & 57.3 & \textbf{49.2} \\
\midrule
\texttt{gold zero-shot} & - & - & - & - & \textbf{81.9} & \textbf{75.9} & \textbf{67.9} & - \\
\texttt{silver zero-shot} & \textbf{90.3} & \textbf{77.0} & 51.9 & \textbf{57.5} & - & - & - & 46.4 \\
\bottomrule
\end{tabular}
\caption{Transfer learning results averaged across \textbf{languages} from the \benchmark/ benchmark. We report F1 scores for Structure Prediction \& QA, and accuracy for the other tasks.}
\label{tab:transfer-results}
\end{table*}

\paragraph{Effect of Monolingual Corpora} Table \ref{tab:results} compares the results for IndicBERT trained on IndicCorp v1 and v2. We can clearly see that model trained on the much larger v2 corpora performs better than model trained with v1 (see \texttt{v1-data} in Table \ref{tab:results}), thereby establishing the utility of the larger monolingual corpora which we release as a part of this work.

\paragraph{Utilizing language similarity} All models in Table \ref{tab:results} are optimized using English development sets. We can get better performance from these models if we have access to in-language development sets. This is not always possible since it may involve expensive and time-consuming human annotations. An alternate approach is to use machine-translated developments sets. For some languages, getting these translations is also impossible. In such cases, we might be able to use a surrogate development set from a different language that has similar linguistic properties. Often, this condition is satisfied by a sibling language from the same family subtree. 

To test this hypothesis, we fine-tune models with in-language development sets if available, and compare their performance with those fine-tuned with in-family development sets. We use Hindi and Tamil development sets to select the best models for Indo-European and Dravidian languages respectively and the results are shown in Table \ref{tab:dev-set-analysis}. We see that models fine-tuned with in-family development sets generally perform on par with those fine-tuned with in-language sets, and give better performance than that obtained using English validation sets.

\paragraph{Shared Script} Prior works \citet{DBLP:journals/tacl/RameshDBJASSDJK22, DBLP:conf/acl/KhemchandaniMPA20} established that having a shared script model helps in lexical sharing leading to better performance. Taking inspiration from this, we train \texttt{IndicBERT-SS}. Largely the performance of \texttt{IndicBERT-SS} is comparable to models without script sharing, however, it does improve the performance of low resource languages written in Devanagari, see Tables \ref{app-tab:xcopa}, \ref{app-tab:flores} in Appendix. 

\paragraph{Transfer Languages} We use English as the  transfer language given the availability of sufficient training data for most tasks, but it might not be the best choice and another similar ``related'' language might be a better transfer language \cite{lauscher-etal-2020-zero,lin-etal-2019-choosing}. We conduct a preliminary experiment to verify this observation on the Naamapadam and MASSIVE datasets for Indic languages (which contains both training and development sets in multiple languages). Here, we compare Hindi (a ``related'' language) with English as the transfer language (Table \ref{tab:transfer-results}, \texttt{gold zero-shot}). We also compare this across models (Table \ref{tab:hi-zero-shot}). For NER we see a significant jump of 3.5 points when fine-tuning with Hindi. Similarly, for MASSIVE we see gains of 3.1 and 10.6 for Intent classification and slot filling respectively. These results suggest that it is useful to leverage training data in a related language. Prior work also suggests that fine-tuning with data translated to the transfer language \cite{Turc2021RevisitingTP} or the target language \cite{DBLP:journals/corr/abs-2204-08776, pmlr-v119-hu20b} (translate-train method) can perform better than when English is used as a transfer language. We plan to do further experiments with more tasks to investigate these observations broadly for Indic language settings. We call upon the community to create and share more in-language data, either through human annotation or (semi-)automated techniques.

\paragraph{\texttt{Silver} zero-shot} To further test the hypothesis that zero-shot with ``related'' language results in better performance, we surrogate the English training data with translated data. Specifically, we translate the English training data for tasks to Hindi (w/ \cite{DBLP:journals/tacl/RameshDBJASSDJK22}) and use this for zero-shot transfer. For QA, we use the translation released by authors of \citet{DBLP:conf/acl/LewisORRS20}. The results are shown in Table \ref{tab:results}. We see that zero-shot with silver translation leads to much better performance than with English. The COPA task is generally described as a much harder task and even small perturbations in the data leads to bad performance. Similarly, translating QA datasets by preserving the answers spans is typically error prone, so we see a slight drop in performance for QA task.

\paragraph{``Winners'' vs. ``Left-Behinds''} Table \ref{tab:lang-wise-results} presents language-wise results which are averaged across tasks. We can see a clear performance drop for extremely low-resource languages (those below the 10th percentile in Table \ref{tab:corpus_stats}). For example, Santhali and Sindhi performance on IndicXCOPA is 25.9\% \& 17.7\% less than that for Hindi. Apart from lacking pretraining data, there are two other important reasons for this drop: (i) no shared script among languages, and (ii) no linguistic cousin in the corpus to act as a bridge for effective transfer. It is to be noted that \benchmark/ can only evaluate 19 of the 24 languages present in \corpus/. There is an urgent need to build datasets for these ``left-behind'' languages.

\section{Conclusion}
Through this work, we distinctively contribute towards all the fundamental requirements of developing Indic language technologies; 
These include \corpus/ v2, the largest pretraining corpus for 24 Indic languages, \model/ v2 a language model pretrained on \corpus/ v2 and a holistic cross-lingual NLU benchmark, \benchmark/, for 20 Indic languages. 
We provide empirical evidence for our design decisions and show that pretraining models only on Indic languages result in much better performance on \benchmark/.

\section*{Acknowledgements}
We would like to thank the Ministry of Electronics and Information Technology\footnote{\url{https://www.meity.gov.in/}} of the Government of India for their generous grant through the Digital India Bhashini project\footnote{\url{https://www.bhashini.gov.in/}}. We also thank the Centre for Development of Advanced Computing\footnote{\url{ https://www.cdac.in/index.aspx?id=pune}} for providing compute time on the Param Siddhi Supercomputer. We also thank Nilekani Philanthropies for their generous grant towards building datasets, models, tools and resources for Indic languages. We also thank Microsoft for their grant to support research on Indic languages.  We also thank Google's TPU Research Cloud (TRC) for giving us free access to their v3-128 TPUs for pretraining our models. We would like to thank Janki Nawale, Anupama Sujatha, and Krishnan Karunganni for their help in coordinating the annotation work. Most importantly we would like to thank all the annotators who spent their time helping create the \benchmark/ benchmark. We also thank Raghavan AK for helpful discussions on corpus cleaning and Harshita Diddee for insightful discussions on model pretraining.

\section*{Limitations}
To create a clean and diverse corpus, we have chosen to crawl news articles as our primary data sources. Since all the articles are crawled from public domains, the data could potentially encompass the biases which propagate in public channels. Currently, the models trained on such data sources could model the inherent biases present within the data. In the current work, we do not perform any debiasing techniques and leave that for future work. 

Language Identification (LID) tools are restricted to a limited number of languages and unavailable for some of the very low-resource languages like Bodo, Dogri, Khasi, etc. We made our best effort to clean the corpus using Unicode spans, but it is possible that the data sources could have some issues. We leave developing LID tools for low-resource languages as part of future work.

From our ablation studies, we see that models are benefited by using in-language training and/or development sets. We call upon the community to work together to create more in-language data resources. Finally, there is still work required in terms of building datasets for hundreds of extremely low-resource languages not represented in this work.

\section*{Ethics Statement}
Annotators who participated in the annotation and/or verification task are paid a competitive monthly salary to help with the tasks. The salaries were determined based on the qualification and the prior experience working on similar tasks and adhering to the norms of the government of our country. All the annotators were native speakers of the respective languages and from the Indian subcontinent. The annotators were made aware that the datasets will be publicly released. The annotated datasets have no personally identifying information.
The annotated data and the crawled corpus have been checked for any offensive data and discarded if present. 

The released code and models will have an MIT License\footnote{\url{https://opensource.org/licenses/MIT}}. The dataset will be released under a CC-0 License\footnote{\url{https://creativecommons.org/share-your-work/public-domain/cc0/}}.

\bibliography{anthology,custom}

\begin{thebibliography}{49}
\expandafter\ifx\csname natexlab\endcsname\relax\def\natexlab#1{#1}\fi

\bibitem[{Aggarwal et~al.(2022)Aggarwal, Gupta, and
  Kunchukuttan}]{DBLP:journals/corr/abs-2204-08776}
Divyanshu Aggarwal, Vivek Gupta, and Anoop Kunchukuttan. 2022.
\newblock \href {https://doi.org/10.48550/arXiv.2204.08776} {Indicxnli:
  Evaluating multilingual inference for indian languages}.
\newblock \emph{CoRR}, abs/2204.08776.

\bibitem[{Conneau et~al.(2020)Conneau, Khandelwal, Goyal, Chaudhary, Wenzek,
  Guzm{\'a}n, Grave, Ott, Zettlemoyer, and
  Stoyanov}]{conneau-etal-2020-unsupervised}
Alexis Conneau, Kartikay Khandelwal, Naman Goyal, Vishrav Chaudhary, Guillaume
  Wenzek, Francisco Guzm{\'a}n, Edouard Grave, Myle Ott, Luke Zettlemoyer, and
  Veselin Stoyanov. 2020.
\newblock \href {https://doi.org/10.18653/v1/2020.acl-main.747} {Unsupervised
  cross-lingual representation learning at scale}.
\newblock In \emph{Proceedings of the 58th Annual Meeting of the Association
  for Computational Linguistics}, pages 8440--8451, Online. Association for
  Computational Linguistics.

\bibitem[{Conneau and Lample(2019)}]{DBLP:conf/nips/ConneauL19}
Alexis Conneau and Guillaume Lample. 2019.
\newblock \href
  {https://proceedings.neurips.cc/paper/2019/hash/c04c19c2c2474dbf5f7ac4372c5b9af1-Abstract.html}
  {Cross-lingual language model pretraining}.
\newblock In \emph{Advances in Neural Information Processing Systems 32: Annual
  Conference on Neural Information Processing Systems 2019, NeurIPS 2019,
  December 8-14, 2019, Vancouver, BC, Canada}, pages 7057--7067.

\bibitem[{Conneau et~al.(2018)Conneau, Rinott, Lample, Williams, Bowman,
  Schwenk, and Stoyanov}]{DBLP:conf/emnlp/ConneauRLWBSS18}
Alexis Conneau, Ruty Rinott, Guillaume Lample, Adina Williams, Samuel~R.
  Bowman, Holger Schwenk, and Veselin Stoyanov. 2018.
\newblock \href {https://doi.org/10.18653/v1/d18-1269} {{XNLI:} evaluating
  cross-lingual sentence representations}.
\newblock In \emph{Proceedings of the 2018 Conference on Empirical Methods in
  Natural Language Processing, Brussels, Belgium, October 31 - November 4,
  2018}, pages 2475--2485. Association for Computational Linguistics.

\bibitem[{Costa{-}juss{\`{a}} et~al.(2022)Costa{-}juss{\`{a}}, Cross,
  {\c{C}}elebi, Elbayad, Heafield, Heffernan, Kalbassi, Lam, Licht, Maillard,
  Sun, Wang, Wenzek, Youngblood, Akula, Barrault, Gonzalez, Hansanti, Hoffman,
  Jarrett, Sadagopan, Rowe, Spruit, Tran, Andrews, Ayan, Bhosale, Edunov, Fan,
  Gao, Goswami, Guzm{\'{a}}n, Koehn, Mourachko, Ropers, Saleem, Schwenk, and
  Wang}]{DBLP:journals/corr/abs-2207-04672}
Marta~R. Costa{-}juss{\`{a}}, James Cross, Onur {\c{C}}elebi, Maha Elbayad,
  Kenneth Heafield, Kevin Heffernan, Elahe Kalbassi, Janice Lam, Daniel Licht,
  Jean Maillard, Anna Sun, Skyler Wang, Guillaume Wenzek, Al~Youngblood, Bapi
  Akula, Lo{\"{\i}}c Barrault, Gabriel~Mejia Gonzalez, Prangthip Hansanti, John
  Hoffman, Semarley Jarrett, Kaushik~Ram Sadagopan, Dirk Rowe, Shannon Spruit,
  Chau Tran, Pierre Andrews, Necip~Fazil Ayan, Shruti Bhosale, Sergey Edunov,
  Angela Fan, Cynthia Gao, Vedanuj Goswami, Francisco Guzm{\'{a}}n, Philipp
  Koehn, Alexandre Mourachko, Christophe Ropers, Safiyyah Saleem, Holger
  Schwenk, and Jeff Wang. 2022.
\newblock \href {https://doi.org/10.48550/arXiv.2207.04672} {No language left
  behind: Scaling human-centered machine translation}.
\newblock \emph{CoRR}, abs/2207.04672.

\bibitem[{Dabre et~al.(2022)Dabre, Shrotriya, Kunchukuttan, Puduppully, Khapra,
  and Kumar}]{dabre-etal-2022-indicbart}
Raj Dabre, Himani Shrotriya, Anoop Kunchukuttan, Ratish Puduppully, Mitesh
  Khapra, and Pratyush Kumar. 2022.
\newblock \href {https://doi.org/10.18653/v1/2022.findings-acl.145}
  {{I}ndic{BART}: A pre-trained model for indic natural language generation}.
\newblock In \emph{Findings of the Association for Computational Linguistics:
  ACL 2022}, pages 1849--1863, Dublin, Ireland. Association for Computational
  Linguistics.

\bibitem[{Devlin et~al.(2019)Devlin, Chang, Lee, and
  Toutanova}]{DBLP:conf/naacl/DevlinCLT19}
Jacob Devlin, Ming{-}Wei Chang, Kenton Lee, and Kristina Toutanova. 2019.
\newblock \href {https://doi.org/10.18653/v1/n19-1423} {{BERT:} pre-training of
  deep bidirectional transformers for language understanding}.
\newblock In \emph{Proceedings of the 2019 Conference of the North American
  Chapter of the Association for Computational Linguistics: Human Language
  Technologies, {NAACL-HLT} 2019, Minneapolis, MN, USA, June 2-7, 2019, Volume
  1 (Long and Short Papers)}, pages 4171--4186. Association for Computational
  Linguistics.

\bibitem[{Doddapaneni et~al.(2021)Doddapaneni, Ramesh, Kunchukuttan, Kumar, and
  Khapra}]{DBLP:journals/corr/abs-2107-00676}
Sumanth Doddapaneni, Gowtham Ramesh, Anoop Kunchukuttan, Pratyush Kumar, and
  Mitesh~M. Khapra. 2021.
\newblock \href {http://arxiv.org/abs/2107.00676} {A primer on pretrained
  multilingual language models}.
\newblock \emph{CoRR}, abs/2107.00676.

\bibitem[{FitzGerald et~al.(2022)FitzGerald, Hench, Peris, Mackie, Rottmann,
  Sanchez, Nash, Urbach, Kakarala, Singh, Ranganath, Crist, Britan, Leeuwis,
  T{\"{u}}r, and Natarajan}]{DBLP:journals/corr/abs-2204-08582}
Jack FitzGerald, Christopher Hench, Charith Peris, Scott Mackie, Kay Rottmann,
  Ana Sanchez, Aaron Nash, Liam Urbach, Vishesh Kakarala, Richa Singh, Swetha
  Ranganath, Laurie Crist, Misha Britan, Wouter Leeuwis, G{\"{o}}khan
  T{\"{u}}r, and Prem Natarajan. 2022.
\newblock \href {https://doi.org/10.48550/arXiv.2204.08582} {{MASSIVE:} {A}
  1m-example multilingual natural language understanding dataset with 51
  typologically-diverse languages}.
\newblock \emph{CoRR}, abs/2204.08582.

\bibitem[{Goyal et~al.(2022)Goyal, Gao, Chaudhary, Chen, Wenzek, Ju, Krishnan,
  Ranzato, Guzm{\'{a}}n, and Fan}]{DBLP:journals/tacl/GoyalGCCWJKRGF22}
Naman Goyal, Cynthia Gao, Vishrav Chaudhary, Peng{-}Jen Chen, Guillaume Wenzek,
  Da~Ju, Sanjana Krishnan, Marc'Aurelio Ranzato, Francisco Guzm{\'{a}}n, and
  Angela Fan. 2022.
\newblock \href {https://doi.org/10.1162/tacl\_a\_00474} {The flores-101
  evaluation benchmark for low-resource and multilingual machine translation}.
\newblock \emph{Trans. Assoc. Comput. Linguistics}, 10:522--538.

\bibitem[{Haddow and Kirefu(2020)}]{DBLP:journals/corr/abs-2001-09907}
Barry Haddow and Faheem Kirefu. 2020.
\newblock \href {http://arxiv.org/abs/2001.09907} {Pmindia - {A} collection of
  parallel corpora of languages of india}.
\newblock \emph{CoRR}, abs/2001.09907.

\bibitem[{Hu et~al.(2020)Hu, Ruder, Siddhant, Neubig, Firat, and
  Johnson}]{pmlr-v119-hu20b}
Junjie Hu, Sebastian Ruder, Aditya Siddhant, Graham Neubig, Orhan Firat, and
  Melvin Johnson. 2020.
\newblock \href {https://proceedings.mlr.press/v119/hu20b.html} {{XTREME}: A
  massively multilingual multi-task benchmark for evaluating cross-lingual
  generalisation}.
\newblock In \emph{Proceedings of the 37th International Conference on Machine
  Learning}, volume 119 of \emph{Proceedings of Machine Learning Research},
  pages 4411--4421. PMLR.

\bibitem[{Joshi et~al.(2020)Joshi, Santy, Budhiraja, Bali, and
  Choudhury}]{joshi-etal-2020-state}
Pratik Joshi, Sebastin Santy, Amar Budhiraja, Kalika Bali, and Monojit
  Choudhury. 2020.
\newblock \href {https://doi.org/10.18653/v1/2020.acl-main.560} {The state and
  fate of linguistic diversity and inclusion in the {NLP} world}.
\newblock In \emph{Proceedings of the 58th Annual Meeting of the Association
  for Computational Linguistics}, pages 6282--6293, Online. Association for
  Computational Linguistics.

\bibitem[{Kakwani et~al.(2020)Kakwani, Kunchukuttan, Golla, N.C.,
  Bhattacharyya, Khapra, and Kumar}]{kakwani-etal-2020-indicnlpsuite}
Divyanshu Kakwani, Anoop Kunchukuttan, Satish Golla, Gokul N.C., Avik
  Bhattacharyya, Mitesh~M. Khapra, and Pratyush Kumar. 2020.
\newblock \href {https://doi.org/10.18653/v1/2020.findings-emnlp.445}
  {{I}ndic{NLPS}uite: Monolingual corpora, evaluation benchmarks and
  pre-trained multilingual language models for {I}ndian languages}.
\newblock In \emph{Findings of the Association for Computational Linguistics:
  EMNLP 2020}, pages 4948--4961, Online. Association for Computational
  Linguistics.

\bibitem[{Keung et~al.(2020)Keung, Lu, Szarvas, and
  Smith}]{DBLP:conf/emnlp/KeungLSS20}
Phillip Keung, Yichao Lu, Gy{\"{o}}rgy Szarvas, and Noah~A. Smith. 2020.
\newblock \href {https://doi.org/10.18653/v1/2020.emnlp-main.369} {The
  multilingual amazon reviews corpus}.
\newblock In \emph{Proceedings of the 2020 Conference on Empirical Methods in
  Natural Language Processing, {EMNLP} 2020, Online, November 16-20, 2020},
  pages 4563--4568. Association for Computational Linguistics.

\bibitem[{Khanuja et~al.(2021)Khanuja, Bansal, Mehtani, Khosla, Dey, Gopalan,
  Margam, Aggarwal, Nagipogu, Dave, Gupta, Gali, Subramanian, and
  Talukdar}]{DBLP:journals/corr/abs-2103-10730}
Simran Khanuja, Diksha Bansal, Sarvesh Mehtani, Savya Khosla, Atreyee Dey,
  Balaji Gopalan, Dilip~Kumar Margam, Pooja Aggarwal, Rajiv~Teja Nagipogu,
  Shachi Dave, Shruti Gupta, Subhash Chandra~Bose Gali, Vish Subramanian, and
  Partha~P. Talukdar. 2021.
\newblock \href {http://arxiv.org/abs/2103.10730} {Muril: Multilingual
  representations for indian languages}.
\newblock \emph{CoRR}, abs/2103.10730.

\bibitem[{Khanuja et~al.(2022)Khanuja, Ruder, and
  Talukdar}]{DBLP:journals/corr/abs-2205-12676}
Simran Khanuja, Sebastian Ruder, and Partha~P. Talukdar. 2022.
\newblock \href {https://doi.org/10.48550/arXiv.2205.12676} {Evaluating
  inclusivity, equity, and accessibility of {NLP} technology: {A} case study
  for indian languages}.
\newblock \emph{CoRR}, abs/2205.12676.

\bibitem[{Khemchandani et~al.(2021)Khemchandani, Mehtani, Patil, Awasthi,
  Talukdar, and Sarawagi}]{DBLP:conf/acl/KhemchandaniMPA20}
Yash Khemchandani, Sarvesh Mehtani, Vaidehi Patil, Abhijeet Awasthi, Partha~P.
  Talukdar, and Sunita Sarawagi. 2021.
\newblock \href {https://doi.org/10.18653/v1/2021.acl-long.105} {Exploiting
  language relatedness for low web-resource language model adaptation: An indic
  languages study}.
\newblock In \emph{Proceedings of the 59th Annual Meeting of the Association
  for Computational Linguistics and the 11th International Joint Conference on
  Natural Language Processing, {ACL/IJCNLP} 2021, (Volume 1: Long Papers),
  Virtual Event, August 1-6, 2021}, pages 1312--1323. Association for
  Computational Linguistics.

\bibitem[{Kreutzer et~al.(2022{\natexlab{a}})Kreutzer, Caswell, Wang, Wahab,
  van Esch, Ulzii{-}Orshikh, Tapo, Subramani, Sokolov, Sikasote, Setyawan,
  Sarin, Samb, Sagot, Rivera, Rios, Papadimitriou, Osei, Su{\'{a}}rez, Orife,
  Ogueji, Rubungo, Nguyen, M{\"{u}}ller, M{\"{u}}ller, Muhammad, Muhammad,
  Mnyakeni, Mirzakhalov, Matangira, Leong, Lawson, Kudugunta, Jernite, Jenny,
  Firat, Dossou, Dlamini, de~Silva, Balli, Biderman, Battisti, Baruwa, Bapna,
  Baljekar, Azime, Awokoya, Ataman, Ahia, Ahia, Agrawal, and
  Adeyemi}]{DBLP:journals/tacl/KreutzerCWWEUTS22}
Julia Kreutzer, Isaac Caswell, Lisa Wang, Ahsan Wahab, Daan van Esch,
  Nasanbayar Ulzii{-}Orshikh, Allahsera Tapo, Nishant Subramani, Artem Sokolov,
  Claytone Sikasote, Monang Setyawan, Supheakmungkol Sarin, Sokhar Samb,
  Beno{\^{\i}}t Sagot, Clara Rivera, Annette Rios, Isabel Papadimitriou,
  Salomey Osei, Pedro Javier~Ortiz Su{\'{a}}rez, Iroro Orife, Kelechi Ogueji,
  Andre~Niyongabo Rubungo, Toan~Q. Nguyen, Mathias M{\"{u}}ller, Andr{\'{e}}
  M{\"{u}}ller, Shamsuddeen~Hassan Muhammad, Nanda Muhammad, Ayanda Mnyakeni,
  Jamshidbek Mirzakhalov, Tapiwanashe Matangira, Colin Leong, Nze Lawson, Sneha
  Kudugunta, Yacine Jernite, Mathias Jenny, Orhan Firat, Bonaventure F.~P.
  Dossou, Sakhile Dlamini, Nisansa de~Silva, Sakine~{\c{C}}abuk Balli, Stella
  Biderman, Alessia Battisti, Ahmed Baruwa, Ankur Bapna, Pallavi Baljekar,
  Israel~Abebe Azime, Ayodele Awokoya, Duygu Ataman, Orevaoghene Ahia,
  Oghenefego Ahia, Sweta Agrawal, and Mofetoluwa Adeyemi. 2022{\natexlab{a}}.
\newblock \href {https://doi.org/10.1162/tacl\_a\_00447} {Quality at a glance:
  An audit of web-crawled multilingual datasets}.
\newblock \emph{Trans. Assoc. Comput. Linguistics}, 10:50--72.

\bibitem[{Kreutzer et~al.(2022{\natexlab{b}})Kreutzer, Caswell, Wang, Wahab,
  van Esch, Ulzii-Orshikh, Tapo, Subramani, Sokolov, Sikasote, Setyawan, Sarin,
  Samb, Sagot, Rivera, Rios, Papadimitriou, Osei, Suarez, Orife, Ogueji,
  Rubungo, Nguyen, M{\"u}ller, M{\"u}ller, Muhammad, Muhammad, Mnyakeni,
  Mirzakhalov, Matangira, Leong, Lawson, Kudugunta, Jernite, Jenny, Firat,
  Dossou, Dlamini, de~Silva, {\c{C}}abuk~Ball{\i}, Biderman, Battisti, Baruwa,
  Bapna, Baljekar, Azime, Awokoya, Ataman, Ahia, Ahia, Agrawal, and
  Adeyemi}]{kreutzer-etal-2022-quality}
Julia Kreutzer, Isaac Caswell, Lisa Wang, Ahsan Wahab, Daan van Esch,
  Nasanbayar Ulzii-Orshikh, Allahsera Tapo, Nishant Subramani, Artem Sokolov,
  Claytone Sikasote, Monang Setyawan, Supheakmungkol Sarin, Sokhar Samb,
  Beno{\^\i}t Sagot, Clara Rivera, Annette Rios, Isabel Papadimitriou, Salomey
  Osei, Pedro~Ortiz Suarez, Iroro Orife, Kelechi Ogueji, Andre~Niyongabo
  Rubungo, Toan~Q. Nguyen, Mathias M{\"u}ller, Andr{\'e} M{\"u}ller,
  Shamsuddeen~Hassan Muhammad, Nanda Muhammad, Ayanda Mnyakeni, Jamshidbek
  Mirzakhalov, Tapiwanashe Matangira, Colin Leong, Nze Lawson, Sneha Kudugunta,
  Yacine Jernite, Mathias Jenny, Orhan Firat, Bonaventure F.~P. Dossou, Sakhile
  Dlamini, Nisansa de~Silva, Sakine {\c{C}}abuk~Ball{\i}, Stella Biderman,
  Alessia Battisti, Ahmed Baruwa, Ankur Bapna, Pallavi Baljekar, Israel~Abebe
  Azime, Ayodele Awokoya, Duygu Ataman, Orevaoghene Ahia, Oghenefego Ahia,
  Sweta Agrawal, and Mofetoluwa Adeyemi. 2022{\natexlab{b}}.
\newblock \href {https://doi.org/10.1162/tacl_a_00447} {Quality at a glance: An
  audit of web-crawled multilingual datasets}.
\newblock \emph{Transactions of the Association for Computational Linguistics},
  10:50--72.

\bibitem[{Kumar et~al.(2022)Kumar, Shrotriya, Sahu, Dabre, Puduppully,
  Kunchukuttan, Mishra, Khapra, and Kumar}]{DBLP:journals/corr/abs-2203-05437}
Aman Kumar, Himani Shrotriya, Prachi Sahu, Raj Dabre, Ratish Puduppully, Anoop
  Kunchukuttan, Amogh Mishra, Mitesh~M. Khapra, and Pratyush Kumar. 2022.
\newblock \href {https://doi.org/10.48550/arXiv.2203.05437} {Indicnlg suite:
  Multilingual datasets for diverse {NLG} tasks in indic languages}.
\newblock \emph{CoRR}, abs/2203.05437.

\bibitem[{Kunchukuttan(2020)}]{kunchukuttan2020indicnlp}
Anoop Kunchukuttan. 2020.
\newblock {The IndicNLP Library}.
\newblock
  \url{https://github.com/anoopkunchukuttan/indic_nlp_library/blob/master/docs/indicnlp.pdf}.

\bibitem[{Lan et~al.(2020)Lan, Chen, Goodman, Gimpel, Sharma, and
  Soricut}]{DBLP:conf/iclr/LanCGGSS20}
Zhenzhong Lan, Mingda Chen, Sebastian Goodman, Kevin Gimpel, Piyush Sharma, and
  Radu Soricut. 2020.
\newblock \href {https://openreview.net/forum?id=H1eA7AEtvS} {{ALBERT:} {A}
  lite {BERT} for self-supervised learning of language representations}.
\newblock In \emph{8th International Conference on Learning Representations,
  {ICLR} 2020, Addis Ababa, Ethiopia, April 26-30, 2020}. OpenReview.net.

\bibitem[{Lauscher et~al.(2020)Lauscher, Ravishankar, Vuli{\'c}, and
  Glava{\v{s}}}]{lauscher-etal-2020-zero}
Anne Lauscher, Vinit Ravishankar, Ivan Vuli{\'c}, and Goran Glava{\v{s}}. 2020.
\newblock \href {https://doi.org/10.18653/v1/2020.emnlp-main.363} {From zero to
  hero: {O}n the limitations of zero-shot language transfer with multilingual
  {T}ransformers}.
\newblock In \emph{Proceedings of the 2020 Conference on Empirical Methods in
  Natural Language Processing (EMNLP)}, pages 4483--4499, Online. Association
  for Computational Linguistics.

\bibitem[{Lewis et~al.(2020)Lewis, Oguz, Rinott, Riedel, and
  Schwenk}]{DBLP:conf/acl/LewisORRS20}
Patrick S.~H. Lewis, Barlas Oguz, Ruty Rinott, Sebastian Riedel, and Holger
  Schwenk. 2020.
\newblock \href {https://doi.org/10.18653/v1/2020.acl-main.653} {{MLQA:}
  evaluating cross-lingual extractive question answering}.
\newblock In \emph{Proceedings of the 58th Annual Meeting of the Association
  for Computational Linguistics, {ACL} 2020, Online, July 5-10, 2020}, pages
  7315--7330. Association for Computational Linguistics.

\bibitem[{Liang et~al.(2020)Liang, Duan, Gong, Wu, Guo, Qi, Gong, Shou, Jiang,
  Cao, Fan, Zhang, Agrawal, Cui, Wei, Bharti, Qiao, Chen, Wu, Liu, Yang,
  Campos, Majumder, and Zhou}]{liang-etal-2020-xglue}
Yaobo Liang, Nan Duan, Yeyun Gong, Ning Wu, Fenfei Guo, Weizhen Qi, Ming Gong,
  Linjun Shou, Daxin Jiang, Guihong Cao, Xiaodong Fan, Ruofei Zhang, Rahul
  Agrawal, Edward Cui, Sining Wei, Taroon Bharti, Ying Qiao, Jiun-Hung Chen,
  Winnie Wu, Shuguang Liu, Fan Yang, Daniel Campos, Rangan Majumder, and Ming
  Zhou. 2020.
\newblock \href {https://doi.org/10.18653/v1/2020.emnlp-main.484} {{XGLUE}: A
  new benchmark datasetfor cross-lingual pre-training, understanding and
  generation}.
\newblock In \emph{Proceedings of the 2020 Conference on Empirical Methods in
  Natural Language Processing (EMNLP)}, pages 6008--6018, Online. Association
  for Computational Linguistics.

\bibitem[{Lin et~al.(2019)Lin, Chen, Lee, Li, Zhang, Xia, Rijhwani, He, Zhang,
  Ma, Anastasopoulos, Littell, and Neubig}]{lin-etal-2019-choosing}
Yu-Hsiang Lin, Chian-Yu Chen, Jean Lee, Zirui Li, Yuyan Zhang, Mengzhou Xia,
  Shruti Rijhwani, Junxian He, Zhisong Zhang, Xuezhe Ma, Antonios
  Anastasopoulos, Patrick Littell, and Graham Neubig. 2019.
\newblock \href {https://doi.org/10.18653/v1/P19-1301} {Choosing transfer
  languages for cross-lingual learning}.
\newblock In \emph{Proceedings of the 57th Annual Meeting of the Association
  for Computational Linguistics}, pages 3125--3135, Florence, Italy.
  Association for Computational Linguistics.

\bibitem[{Liu et~al.(2019)Liu, Ott, Goyal, Du, Joshi, Chen, Levy, Lewis,
  Zettlemoyer, and Stoyanov}]{DBLP:journals/corr/abs-1907-11692}
Yinhan Liu, Myle Ott, Naman Goyal, Jingfei Du, Mandar Joshi, Danqi Chen, Omer
  Levy, Mike Lewis, Luke Zettlemoyer, and Veselin Stoyanov. 2019.
\newblock \href {http://arxiv.org/abs/1907.11692} {Roberta: {A} robustly
  optimized {BERT} pretraining approach}.
\newblock \emph{CoRR}, abs/1907.11692.

\bibitem[{Loshchilov and Hutter(2019)}]{loshchilov2018decoupled}
Ilya Loshchilov and Frank Hutter. 2019.
\newblock \href {https://openreview.net/forum?id=Bkg6RiCqY7} {Decoupled weight
  decay regularization}.
\newblock In \emph{International Conference on Learning Representations}.

\bibitem[{Mhaske et~al.(2022)Mhaske, Kedia, V, Kunchukuttan, Kumar, and
  Khapra}]{mhaske2022indicner}
Arnav Mhaske, Harshit Kedia, Rudramurthy. V, Anoop Kunchukuttan, Pratyush
  Kumar, and Mitesh~M. Khapra. 2022.
\newblock \href {https://huggingface.co/datasets/ai4bharat/naamapadam/}
  {Naamapadam: A large-scale named entity annotated data for indic languages}.

\bibitem[{Ponti et~al.(2021)Ponti, Aralikatte, Shrivastava, Reddy, and
  S{\o}gaard}]{ponti-etal-2021-minimax}
Edoardo~Maria Ponti, Rahul Aralikatte, Disha Shrivastava, Siva Reddy, and
  Anders S{\o}gaard. 2021.
\newblock \href {https://doi.org/10.18653/v1/2021.findings-acl.106} {Minimax
  and neyman{--}{P}earson meta-learning for outlier languages}.
\newblock In \emph{Findings of the Association for Computational Linguistics:
  ACL-IJCNLP 2021}, pages 1245--1260, Online. Association for Computational
  Linguistics.

\bibitem[{Rajpurkar et~al.(2016)Rajpurkar, Zhang, Lopyrev, and
  Liang}]{rajpurkar-etal-2016-squad}
Pranav Rajpurkar, Jian Zhang, Konstantin Lopyrev, and Percy Liang. 2016.
\newblock \href {https://doi.org/10.18653/v1/D16-1264} {{SQ}u{AD}: 100,000+
  questions for machine comprehension of text}.
\newblock In \emph{Proceedings of the 2016 Conference on Empirical Methods in
  Natural Language Processing}, pages 2383--2392, Austin, Texas. Association
  for Computational Linguistics.

\bibitem[{Ramesh et~al.(2022)Ramesh, Doddapaneni, Bheemaraj, Jobanputra, AK,
  Sharma, Sahoo, Diddee, J, Kakwani, Kumar, Pradeep, Nagaraj, Kumar, Raghavan,
  Kunchukuttan, Kumar, and Khapra}]{DBLP:journals/tacl/RameshDBJASSDJK22}
Gowtham Ramesh, Sumanth Doddapaneni, Aravinth Bheemaraj, Mayank Jobanputra,
  Raghavan AK, Ajitesh Sharma, Sujit Sahoo, Harshita Diddee, Mahalakshmi J,
  Divyanshu Kakwani, Navneet Kumar, Aswin Pradeep, Srihari Nagaraj, Deepak
  Kumar, Vivek Raghavan, Anoop Kunchukuttan, Pratyush Kumar, and
  Mitesh~Shantadevi Khapra. 2022.
\newblock \href {https://doi.org/10.1162/tacl\_a\_00452} {Samanantar: The
  largest publicly available parallel corpora collection for 11 indic
  languages}.
\newblock \emph{Trans. Assoc. Comput. Linguistics}, 10:145--162.

\bibitem[{Reid et~al.(2021)Reid, Hu, Neubig, and
  Matsuo}]{reid-etal-2021-afromt}
Machel Reid, Junjie Hu, Graham Neubig, and Yutaka Matsuo. 2021.
\newblock \href {https://doi.org/10.18653/v1/2021.emnlp-main.99} {{A}fro{MT}:
  Pretraining strategies and reproducible benchmarks for translation of 8
  {A}frican languages}.
\newblock In \emph{Proceedings of the 2021 Conference on Empirical Methods in
  Natural Language Processing}, pages 1306--1320, Online and Punta Cana,
  Dominican Republic. Association for Computational Linguistics.

\bibitem[{Roark et~al.(2020)Roark, Wolf-Sonkin, Kirov, Mielke, Johny,
  Demirsahin, and Hall}]{roark-etal-2020-processing}
Brian Roark, Lawrence Wolf-Sonkin, Christo Kirov, Sabrina~J. Mielke, Cibu
  Johny, Isin Demirsahin, and Keith Hall. 2020.
\newblock \href {https://aclanthology.org/2020.lrec-1.294} {Processing {S}outh
  {A}sian languages written in the {L}atin script: the {D}akshina dataset}.
\newblock In \emph{Proceedings of the Twelfth Language Resources and Evaluation
  Conference}, pages 2413--2423, Marseille, France. European Language Resources
  Association.

\bibitem[{Roemmele et~al.(2011)Roemmele, Bejan, and
  Gordon}]{roemmele2011choice}
Melissa Roemmele, Cosmin~Adrian Bejan, and Andrew~S Gordon. 2011.
\newblock \href
  {https://people.ict.usc.edu/~gordon/publications/AAAI-SPRING11A.PDF} {Choice
  of plausible alternatives: An evaluation of commonsense causal reasoning}.
\newblock In \emph{2011 AAAI Spring Symposium Series}.

\bibitem[{Ruder et~al.(2021)Ruder, Constant, Botha, Siddhant, Firat, Fu, Liu,
  Hu, Garrette, Neubig, and Johnson}]{DBLP:conf/emnlp/RuderCBSFF00GNJ21}
Sebastian Ruder, Noah Constant, Jan~A. Botha, Aditya Siddhant, Orhan Firat,
  Jinlan Fu, Pengfei Liu, Junjie Hu, Dan Garrette, Graham Neubig, and Melvin
  Johnson. 2021.
\newblock \href {https://doi.org/10.18653/v1/2021.emnlp-main.802} {{XTREME-R:}
  towards more challenging and nuanced multilingual evaluation}.
\newblock In \emph{Proceedings of the 2021 Conference on Empirical Methods in
  Natural Language Processing, {EMNLP} 2021, Virtual Event / Punta Cana,
  Dominican Republic, 7-11 November, 2021}, pages 10215--10245. Association for
  Computational Linguistics.

\bibitem[{Sap et~al.(2019)Sap, Rashkin, Chen, Le~Bras, and
  Choi}]{sap-etal-2019-social}
Maarten Sap, Hannah Rashkin, Derek Chen, Ronan Le~Bras, and Yejin Choi. 2019.
\newblock \href {https://doi.org/10.18653/v1/D19-1454} {Social {IQ}a:
  Commonsense reasoning about social interactions}.
\newblock In \emph{Proceedings of the 2019 Conference on Empirical Methods in
  Natural Language Processing and the 9th International Joint Conference on
  Natural Language Processing (EMNLP-IJCNLP)}, pages 4463--4473, Hong Kong,
  China. Association for Computational Linguistics.

\bibitem[{Suarez et~al.(2019)Suarez, Sagot, and
  Romary}]{OrtizSuarez2019AsynchronousPF}
Pedro~Ortiz Suarez, Beno{\^i}t Sagot, and Laurent Romary. 2019.
\newblock Asynchronous pipeline for processing huge corpora on medium to low
  resource infrastructures.

\bibitem[{Tjong Kim~Sang and
  De~Meulder(2003)}]{tjong-kim-sang-de-meulder-2003-introduction}
Erik~F. Tjong Kim~Sang and Fien De~Meulder. 2003.
\newblock \href {https://www.aclweb.org/anthology/W03-0419} {Introduction to
  the {C}o{NLL}-2003 shared task: Language-independent named entity
  recognition}.
\newblock In \emph{Proceedings of the Seventh Conference on Natural Language
  Learning at {HLT}-{NAACL} 2003}, pages 142--147.

\bibitem[{Turc et~al.(2021)Turc, Lee, Eisenstein, Chang, and
  Toutanova}]{Turc2021RevisitingTP}
Iulia Turc, Kenton Lee, Jacob Eisenstein, Ming-Wei Chang, and Kristina
  Toutanova. 2021.
\newblock Revisiting the primacy of english in zero-shot cross-lingual
  transfer.
\newblock \emph{ArXiv}, abs/2106.16171.

\bibitem[{Vanmassenhove et~al.(2021)Vanmassenhove, Shterionov, and
  Gwilliam}]{Vanmassenhove2021MachineTE}
Eva Vanmassenhove, D.~Shterionov, and M.~Gwilliam. 2021.
\newblock Machine translationese: Effects of algorithmic bias on linguistic
  complexity in machine translation.
\newblock In \emph{EACL}.

\bibitem[{Wang et~al.(2019)Wang, Pruksachatkun, Nangia, Singh, Michael, Hill,
  Levy, and Bowman}]{DBLP:conf/nips/WangPNSMHLB19}
Alex Wang, Yada Pruksachatkun, Nikita Nangia, Amanpreet Singh, Julian Michael,
  Felix Hill, Omer Levy, and Samuel~R. Bowman. 2019.
\newblock \href
  {https://proceedings.neurips.cc/paper/2019/hash/4496bf24afe7fab6f046bf4923da8de6-Abstract.html}
  {Superglue: {A} stickier benchmark for general-purpose language understanding
  systems}.
\newblock In \emph{Advances in Neural Information Processing Systems 32: Annual
  Conference on Neural Information Processing Systems 2019, NeurIPS 2019,
  December 8-14, 2019, Vancouver, BC, Canada}, pages 3261--3275.

\bibitem[{Wang et~al.(2018)Wang, Singh, Michael, Hill, Levy, and
  Bowman}]{wang-etal-2018-glue}
Alex Wang, Amanpreet Singh, Julian Michael, Felix Hill, Omer Levy, and Samuel
  Bowman. 2018.
\newblock \href {https://doi.org/10.18653/v1/W18-5446} {{GLUE}: A multi-task
  benchmark and analysis platform for natural language understanding}.
\newblock In \emph{Proceedings of the 2018 {EMNLP} Workshop {B}lackbox{NLP}:
  Analyzing and Interpreting Neural Networks for {NLP}}, pages 353--355,
  Brussels, Belgium. Association for Computational Linguistics.

\bibitem[{Williams et~al.(2018)Williams, Nangia, and
  Bowman}]{williams-etal-2018-broad}
Adina Williams, Nikita Nangia, and Samuel Bowman. 2018.
\newblock \href {https://doi.org/10.18653/v1/N18-1101} {A broad-coverage
  challenge corpus for sentence understanding through inference}.
\newblock In \emph{Proceedings of the 2018 Conference of the North {A}merican
  Chapter of the Association for Computational Linguistics: Human Language
  Technologies, Volume 1 (Long Papers)}, pages 1112--1122, New Orleans,
  Louisiana. Association for Computational Linguistics.

\bibitem[{Wu et~al.(2016)Wu, Schuster, Chen, Le, Norouzi, Macherey, Krikun,
  Cao, Gao, Macherey, Klingner, Shah, Johnson, Liu, Kaiser, Gouws, Kato, Kudo,
  Kazawa, Stevens, Kurian, Patil, Wang, Young, Smith, Riesa, Rudnick, Vinyals,
  Corrado, Hughes, and Dean}]{DBLP:journals/corr/WuSCLNMKCGMKSJL16}
Yonghui Wu, Mike Schuster, Zhifeng Chen, Quoc~V. Le, Mohammad Norouzi, Wolfgang
  Macherey, Maxim Krikun, Yuan Cao, Qin Gao, Klaus Macherey, Jeff Klingner,
  Apurva Shah, Melvin Johnson, Xiaobing Liu, Lukasz Kaiser, Stephan Gouws,
  Yoshikiyo Kato, Taku Kudo, Hideto Kazawa, Keith Stevens, George Kurian,
  Nishant Patil, Wei Wang, Cliff Young, Jason Smith, Jason Riesa, Alex Rudnick,
  Oriol Vinyals, Greg Corrado, Macduff Hughes, and Jeffrey Dean. 2016.
\newblock \href {http://arxiv.org/abs/1609.08144} {Google's neural machine
  translation system: Bridging the gap between human and machine translation}.
\newblock \emph{CoRR}, abs/1609.08144.

\bibitem[{Xue et~al.(2021)Xue, Constant, Roberts, Kale, Al-Rfou, Siddhant,
  Barua, and Raffel}]{xue-etal-2021-mt5}
Linting Xue, Noah Constant, Adam Roberts, Mihir Kale, Rami Al-Rfou, Aditya
  Siddhant, Aditya Barua, and Colin Raffel. 2021.
\newblock \href {https://doi.org/10.18653/v1/2021.naacl-main.41} {m{T}5: A
  massively multilingual pre-trained text-to-text transformer}.
\newblock In \emph{Proceedings of the 2021 Conference of the North American
  Chapter of the Association for Computational Linguistics: Human Language
  Technologies}, pages 483--498, Online. Association for Computational
  Linguistics.

\bibitem[{Yang et~al.(2019)Yang, Zhang, Tar, and
  Baldridge}]{yang-etal-2019-paws}
Yinfei Yang, Yuan Zhang, Chris Tar, and Jason Baldridge. 2019.
\newblock \href {https://doi.org/10.18653/v1/D19-1382} {{PAWS}-{X}: A
  cross-lingual adversarial dataset for paraphrase identification}.
\newblock In \emph{Proceedings of the 2019 Conference on Empirical Methods in
  Natural Language Processing and the 9th International Joint Conference on
  Natural Language Processing (EMNLP-IJCNLP)}, pages 3687--3692, Hong Kong,
  China. Association for Computational Linguistics.

\bibitem[{Ács(2019)}]{acs2019exploring}
Judit Ács. 2019.
\newblock \href
  {http://juditacs.github.io/2019/02/19/bert-tokenization-stats.html}
  {Exploring bert's vocabulary}.

\end{thebibliography}
\bibliographystyle{acl_natbib}

\clearpage

\appendix

\section{Environmental Impact}
\label{sec:env-impact}
\model/ and its variants are trained on 20.9 billion tokens encompassing 24 Indic languages. The models are trained on v3-128 TPUs.\footnote{The TPUs reside in the Google Cloud Platform which is carbon neutral: \url{https://cloud.google.com/ sustainability}} 
Each model takes 11 days to complete 1 million training steps and we estimate it to consume 9,562.9 kWh of energy with a carbon footprint of 5.4 MTCO2e. All models are further fine-tuned before downstream evaluation. These experiments are carried out on NVIDIA A100 GPUs and we estimate a total usage of 72 kWh of energy which is equivalent to 41.04 kg of CO2e. To limit the pretraining of such models from scratch, and to enable further research, we release all models trained as part of this work. 

\section{Data Distribution}
\label{apx:data-dist}
\begin{figure}[h]
    \centering
    \includegraphics[width=\columnwidth]{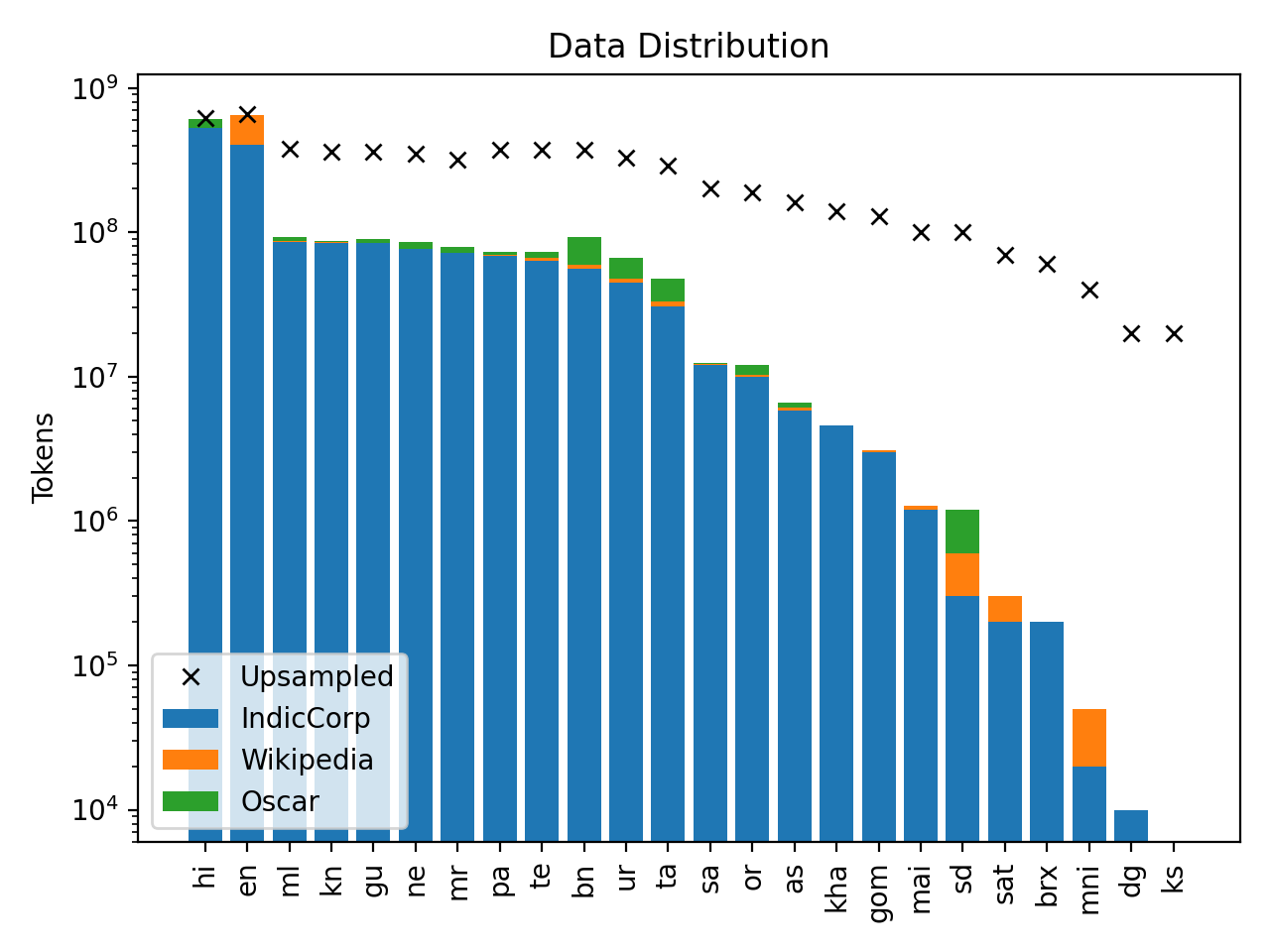}
    \caption{Upsampled data distribution.}
    \label{fig:upsampling}
\end{figure}

\section{Pretraining Hyperparameters}
\label{app:pret-hparams}
% \paragraph{Hyperparameters}
We use the default hyperparameters of BERT-Base with 12 encoder layers, and a maximum sequence length of 512. With 12 attention heads, a hidden dimension of 768, and feedforward network width of 3072, the model has 278 million parameters. We use the AdamW optimizer \cite{loshchilov2018decoupled} with $\alpha=0.9$ and $\beta=0.999$. We use an initial learning rate of 5e-4 with a warm-up of 50,000 steps and linearly decay the learning rate till we reach the 1M steps. We use a global batch size of 4096 examples and train the model on v3-128 TPUs. The models take 11 days to train. More details about environmental impact can be found in Appendix \ref{sec:env-impact}.

\section{Baseline language models}
\label{app:baseline-models}
\paragraph{mBERT} \citep{DBLP:conf/naacl/DevlinCLT19} is one of the first massively multilingual models trained on 104 languages (11 Indic). It is trained on Wikipedia with exponentially smoothed weighting to rectify corpus imbalance. The model has 12 encoder layers with 768-dimensional embeddings and is trained with the MLM objective. It has a vocabulary size of 119,000 and 172 million parameters.

\paragraph{XLM-R} \citep{conneau-etal-2020-unsupervised} is the multilingual version of RoBERTa \cite{DBLP:journals/corr/abs-1907-11692} that is trained on the CC-100 dataset with 100 languages (15 Indic). The model has the same architecture as BERT but has an optimized hyperparameter set. It drops the next-sentence prediction (NSP) objective from the original BERT implementation and uses a combination of MLM and TLM objectives for training. It has a vocabulary size of 250,000, and 278 million parameters.

\paragraph{IndicBERT v1} \citep{kakwani-etal-2020-indicnlpsuite} is a multilingual ALBERT \cite{DBLP:conf/iclr/LanCGGSS20} model trained on IndicCorp v1. The model supports 11 Indic languages. It is smaller than most multilingual models with 33 million parameters.\footnote{Given its small size, we do not perform extensive ablations on this model.} It has a vocabulary size of 200,000 and uses temperature sampling to balance the data across languages. It is trained with the MLM objective, a smaller maximum sequence length of 128, and on sentences instead of the standard practice of training on whole documents.

\paragraph{MuRIL} \cite{DBLP:journals/corr/abs-2103-10730} is a multilingual BERT model trained exclusively on 16 Indic languages, with data taken from Wikipedia, OSCAR, PMI corpus \cite{DBLP:journals/corr/abs-2001-09907}, and the Dakshina dataset \cite{roark-etal-2020-processing}. While it follows standard hyperparameter settings and corpus balancing tricks, it stands out by using silver-translated and transliterated data, along with their gold counterparts. It has a vocabulary of 197,000 tokens, 237 million parameters and is trained with both MLM and TLM objectives.

\section{IndicQA}
\label{apx:indicqa}
\begin{table}[h!]
\centering
\small
\setlength{\tabcolsep}{5pt}
% \resizebox{\columnwidth}{!}{%
\begin{tabular}{r|ccc|r|ccc}
\toprule
L. & Q    & A    & NA   & L.  & Q    & A    & NA   \\
\midrule
as  & 1789 & 1225 & 564  &      &      &      &      \\
bn  & 1763 & 1263 & 500  & mr   & 1604 & 1108 & 496  \\
gu  & 2017 & 1273 & 744  & or   & 1680 & 1279 & 401  \\
hi  & 1547 & 1052 & 495  & pa   & 1542 & 1181 & 361  \\
kn  & 1517 & 1138 & 379  & ta   & 1804 & 1276 & 527  \\
ml  & 1589 & 1101 & 488  & te   & 1734 & 1398 & 336  \\ 
\midrule
\multicolumn{5}{r|}{{\bf Total}} & {\bf 18579} & {\bf 13283} & {\bf 5292} \\
\bottomrule
\end{tabular}
% }
\caption{IndicQA statistics. \textbf{Q}: number of questions, \textbf{A}: number of answerable questions, \textbf{NA}: number of unanswerable questions.}
\label{tab:indicqa_stats}
\end{table}

\subsection{Article Selection}
A list of topics related to Indic history, monuments, authors, politicians, festivals, etc., was manually collected. The topics were then ranked by the number of Indic language Wikipedias they appeared in after discarding those that had less than 10 sentences (on average) in their articles. Finally, the articles of the top-ranking topics were used to create the QA pairs.

\subsection{Annotation Process}
From the shortlisted articles, paragraphs containing 8-10 sentences were used as context.\footnote{Smaller paragraphs were merged.} Previous works have shown that annotators often create questions that have a high lexical overlap with the context paragraphs. To avoid this, we divide the collection process into two phases. 

In Phase one, each context is first split into two parts where the second part is smaller, usually containing 2-3 sentences. Both these context paragraphs are then translated into English with Google Translate \footnote{\url{http://translate.google.com}}. The annotators are asked to create questions (in an Indic language) from these translated context paragraphs. This intermediate translation step ensures that the lexical overlap is reduced since the annotators cannot copy a sentence and turn it into a question by prepending a \textit{wh} word.

In Phase two, the first part of the original context paragraph (in an Indic language) is presented to a different annotator and is asked to mark the answer spans for the questions created previously. Since the second part of the context is not provided, the questions created from them become unanswerable. 

On average there were 2-3 annotators per language and all the annotations were done on Haystack tool\footnote{\url{https://github.com/deepset-ai/haystack}}.

\subsection{Annotation Guidelines}
\label{app:indicqa_guidelines}
The annotators were given a set of detailed guidelines to avoid problems seen in previous QA datasets. The list of guidelines for question creation is as follows:
(i) Create a minimum of two questions from each paragraph,
(ii) The answers should not have a span of more than five continuous words,
(iii) The questions should be unambiguous and understandable even if the context is not provided,
(iv) Try to minimize phrase overlapping between the context paragraph and question, and
(v) Create questions in such a way that the answer span is contained within a single sentence of the paragraph.

\noindent The list of guidelines for answer marking is as follows:
(i) The answer should always be a continuous span whose length is not more than five words,
(ii) An entire sentence cannot be marked as an answer,
(iii) The answer cannot be a pronoun, and
(iv) If the context paragraph contains multiple occurrences of the answer string, always mark the one which is most relevant to the question.

\section{IndicXParaphrase}
\label{app:indic_paraphrase}

We randomly choose 1001 English sentences from the dataset introduced in \citet{DBLP:journals/corr/abs-2203-05437}, such that each sentence is at least 10 words long. Next, we machine-translate these sentences into the required languages using the IndicTrans model. Following this, we ask annotators across languages to (i) verify and correct the translations, if required, and (ii) create one paraphrase and a non-paraphrase for each sentence. The instructions to the annotators are as follows: (i) minimize word overlap between the sentence and the paraphrase, (ii) use temporal phrase swapping where ever possible, e.g., \textit{he fell and got hurt} $\rightarrow$ \textit{he got hurt when he fell} (iii) swap active and passive voice, (iv) use synonyms liberally. 

For creating sentences that are not paraphrased, the annotators are instructed to swap named entities, pronouns, adjectives, adverbs, etc. where possible. An example for named entity swapping: \textit{John} drove \textit{Jane} to the market $\rightarrow$ \textit{Jane} drove \textit{John} to the market. They are also instructed to restrict the use of negation and antonyms unless necessary. 
There were 2 annotator per language and the whole task has been carried out on Google Sheets.
% and the annotators are compensated with INR\sd{XX} for every pair of annotation.  

\section{IndicSentiment}
\label{app:sentiment}
We curate a list of products from 15 popular categories from online marketplaces like Amazon\footnote{\url{https://amazon.in}}, Flipkart\footnote{\url{https://flipkart.com}} and Myntra\footnote{\url{https://myntra.com}}. 
% to extract a broach class of products. We the short list 15 broad categories of everyday products ranging from \textit{Entertainment, Health/Wellness, Food, etc}. We further classify these categories based on the range of the products they contain. For instance, within transportation we include sub-categories like \textit{roadways, airways, railways etc}. We then go on find the appropriate products and the service providers/brands for each of the final products. We made efforts to include only products that fall into general usage and hence we restrict ourselves to these 15 broad classes. 
For each product, we first ask annotators to list aspects of the product that they deem important. We then ask a different set of annotators to write reviews for the products, based on the aspects provided in the previous step. We encourage annotators to be natural and draw from their experiences of using the same, or a similar product. We instruct annotators not to use offensive language in the reviews. For example, for the product category {\it dress}, we ask the annotators to write both positive and negative reviews by concentrating on one or more of the following aspects: \textit{material}, \textit{color}, and \textit{sleeves}. The reviews are initially written in English and then manually translated into other languages.
There were 2 annotator per language and the whole task has been carried out on Google Sheets.
% The entire task has been carried out on Google Sheets and the annotators are compensated with INR\sd{XX} for every review and translation.

\section{Naamapadam}
\label{app:ner}

Results for NER task using Hindi data from Naamdapadam. We perform ablations comparing zero-shot transfer via English and Hindi.

\begin{table}[h!]
\small
\centering
\begin{tabular}{lcc}
\toprule
 & en & hi \\
 \midrule
mBERT & 63.0 & 69.4 \\
XLMR & 71.7 & 74.4 \\
MuRIL & 74.3 & 76.2 \\
\midrule
IndicBERT & 73.2 & 76.2 \\
$\quad$ +Samanantar & 72.4 & 75.9 \\
$\quad\quad$ +Back-Trans. & 71.9 & 75.8 \\
 \bottomrule
\end{tabular}
\caption{Naamapadam ``transfer-language" experiment. We restrict the size of the Hindi fine-tuning set to 11k examples to match the size of the English set. We remove English and Hindi testsets while computing the average to avoid skewing the averages.}
\label{tab:hi-zero-shot}
\end{table}

\section{IndicXNLI}
\label{app:xnli-cleaning}
Our effort to manually correct all the translations in the IndicXNLI \cite{DBLP:journals/corr/abs-2204-08776} dataset is currently ongoing. Table \ref{app-tab:xnli-cleaning} \& Table \ref{app-tab:temp-xnli-scores} shows the current status of the project \& current scores, respectively, across all 11 Indic languages. Once the complete test set is verified and cleaned, we plan to update \benchmark/ with the additional data.

\begin{table}[h!]
\setlength{\tabcolsep}{5pt}
\centering
\begin{tabular}{rll|rll}
\toprule
\textbf{Lang.} & \textbf{Ver.} & \textbf{Corr.} & \textbf{Lang.} & \textbf{Ver.} & \textbf{Corr.} \\
\midrule
as  & 3000 & 1918 & mr  & 1648 & 944 \\
bn  & 1510 & 835  & or  & 2107 & 1820 \\
gu  & - & -       & pa  & - & - \\
hi  & 4000 & 1142 & ta  & - & - \\
kn  & 1370 & 264  & te  & 872 & 527 \\
ml  & 3200 & 2427 \\ 
\bottomrule
\end{tabular}
\caption{Of the 5010 test instances in each language, the number of instances verified and corrected so far is presented in the \textbf{Ver.} and \textbf{Corr.} columns respectively.}
\label{app-tab:xnli-cleaning}
\end{table}

\begin{table}[h!]
\setlength{\tabcolsep}{5pt}
\centering
\begin{tabular}{rll|rll}
\toprule
\textbf{Lang.} & \textbf{Org.} & \textbf{HV$^*$} & \textbf{Lang.} & \textbf{Org.} & \textbf{HV$^*$} \\
\midrule
as  & 71.6 & 72.0 & mr  & 73.2 & 73.5 \\
bn  & 76.3 & 76.5  & or  & 74.0 & 73.5 \\
gu  & 75.6 & 75.6       & pa  & 77.2 & 77.7 \\
hi  & 77.5 & 77.5 & ta  & 74.5 & 74.5 \\
kn  & 74.7 & 74.7  & te  & 75.2 & 75.0 \\
ml  & 74.9 & 73.7 \\
\midrule
    &      &      & Avg. & 75.0 & 75.0 \\
\bottomrule
\end{tabular}
\caption{Scores for \texttt{IndicBERT+Samanantar} model on the IndicXNLI proposed by \citet{DBLP:journals/corr/abs-2204-08776} (Org.) \& current state of verified dataset (HV$^*$)}
\label{app-tab:temp-xnli-scores}
\end{table}

\section{IndicCorp Data Cleaning}
\label{app:data_cleaning}
Since most of our data come from Indic news websites, we discover source URLs through online newspaper directories (e.g., w3newspaper\footnote{\url{https://www.w3newspapers.com/}}) and through automated web searches using hand-picked terms in various languages. We manually identify spam websites from the list of sources and remove them.  

\paragraph{Language Identification} We use cld3\footnote{\url{https://github.com/google/cld3}} and langdetect\footnote{\url{https://github.com/shuyo/language-detection}} to detect the language of an article. We use both in parallel since cld3 does not identify Assamese and Oriya.

\paragraph{Script-based cleaning} Often sentences contain transliterations and phrases in other languages, especially English. Therefore, we use Unicode-based verification to determine if sentences are in their native script. We remove a sentence from the corpus if the ratio of the number of characters in the native script to the total number of characters is less than 0.75.

\paragraph{Punctuation-based cleaning} We strip punctuation from sentences and if the length of the stripped document is less than 10 words, then we remove the document from the corpus.

\paragraph{Offensive word filtering} We collect an exhaustive list of offensive words/phrases from online sources, and native speakers.\footnote{The words/phrases obtained from online sources were manually verified by native speakers.} On average, we curated close to 90 words/phrases per language. When suggested by native speakers, we also add ambiguous words to the list, which are not offensive on their own but can be used in offensive contexts.

Sentences containing at least one word from the list are removed from the corpus. In the case of offensive phrases, we remove a sentence only if the whole phrase appears in the sentence.

\section{Tokenizers}
\label{app:tokenizers}
\begin{figure*}
    \centering
    % \rule{4cm}{4cm}
    \includegraphics[width=\textwidth]{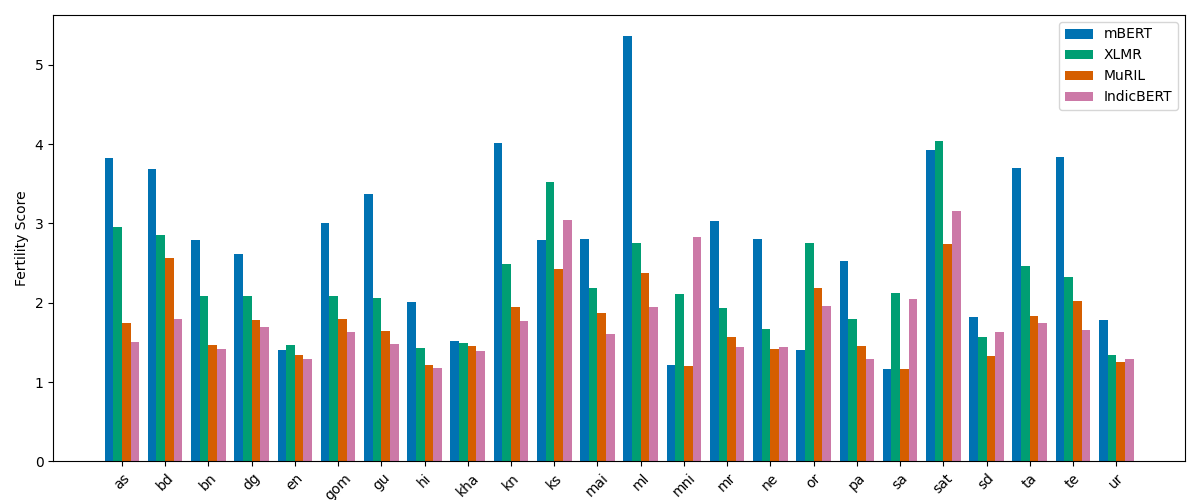}
    \caption{Fertility plots across different tokenizers.}
    \label{fig:fertility}
\end{figure*}

Fig. \ref{fig:fertility} compares the fertility scores \cite{acs2019exploring} of the IndicBERT tokenizer with that of mBERT, XLM-R, and MuRIL. We see that the IndicBERT tokenizer has consistently lower fertility scores across languages which suggests that its vocabulary contains a larger fraction of tokenized words that do not need to be split into subwords. Fertility ratio is higher for \texttt{mni} due to script mismatch between FLORES (Bengali) and IndicCorp (Meitei).

\begin{table}[]
\centering
\begin{tabular}{rc|rc}
\toprule
Lang. & Acc. & Lang. & Acc. \\
\midrule
as       & 100      & mni      & 0        \\
bn       & 100      & ne       & 99.8     \\
gu       & 99.7     & or       & 99.7     \\
hi       & 99.2     & pa       & 99.6     \\
kn       & 100      & sa       & 99.6     \\
ks       & 93.5     & sat      & 99.3    \\
mai      & 99.3     & ta       & 100      \\
ml       & 100      & te       & 100      \\
mr       & 97.1     & ur       & 100     \\
\bottomrule
\end{tabular}
\caption{Language identification results. \texttt{mni} is 0 due to script mismatch between FLORES and IndicCorp.}
\label{tab:lang-id}
\end{table}

\section{Language Identification}
\label{app:lang_id}
Since IndicBERT is pretrained with prepended \texttt{<lang-id>} tags, we evaluate its language identification ability without any fine-tuning. We use the FLORES \textit{devtest} split for this evaluation. We pass the input sentences by prepending the [MASK] token and expect the model to replace it with the appropriate \texttt{<lang-id>}. For this experiment, we only consider top-1 accuracy. See Table \ref{tab:lang-id} for results. Apart from Manipuri, IndicBERT identifies all other languages with high accuracy. It cannot identify Manipuri since FLORES uses the Bengali script for Manipuri, whereas IndicCorp uses Meitei.

\begin{figure*}
     \centering
     \begin{subfigure}[b]{0.3\textwidth}
         \centering
         \includegraphics[width=\textwidth]{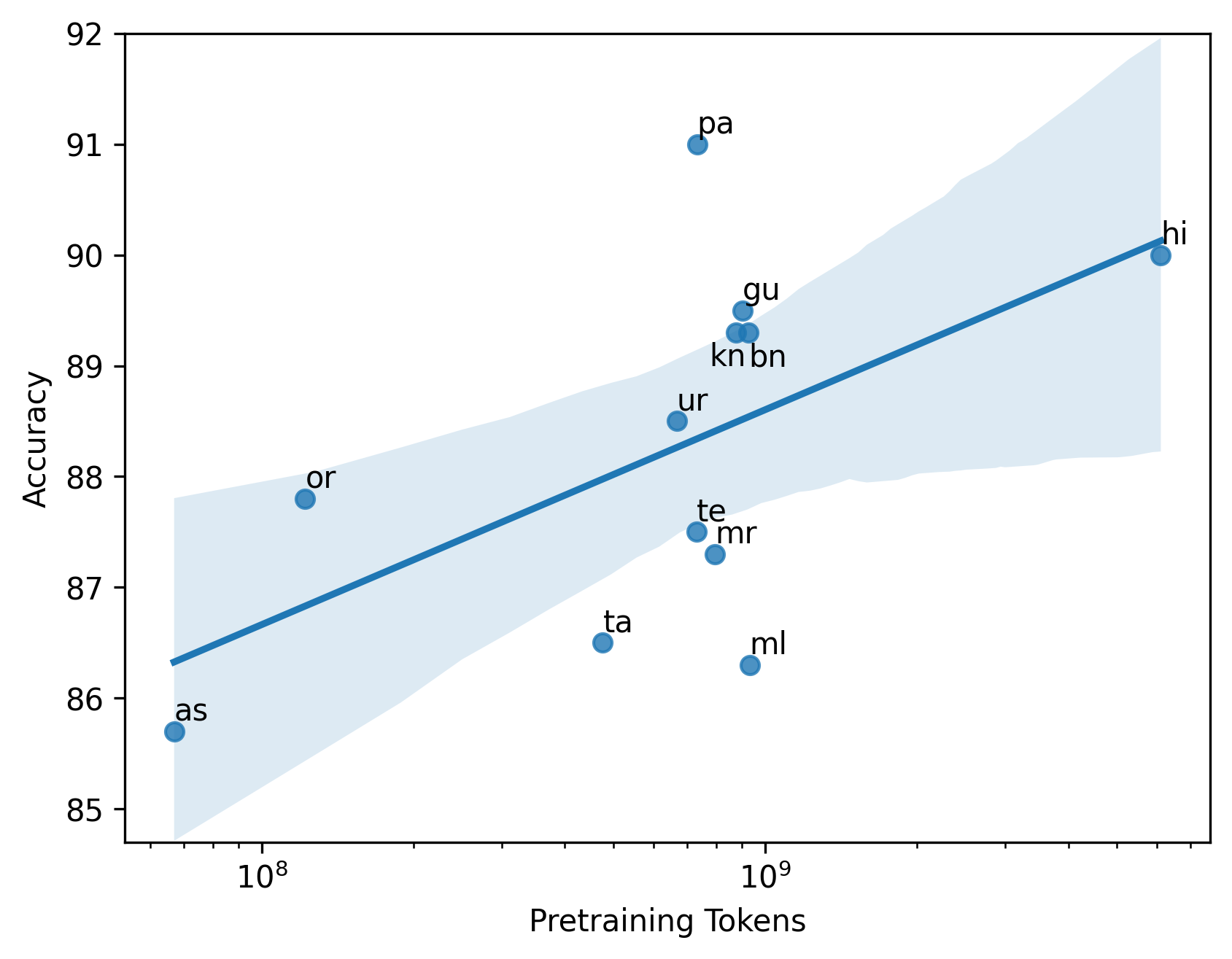}
         \caption{IndicSentiment}
     \end{subfigure}
     \hfill
     \begin{subfigure}[b]{0.3\textwidth}
         \centering
         \includegraphics[width=\textwidth]{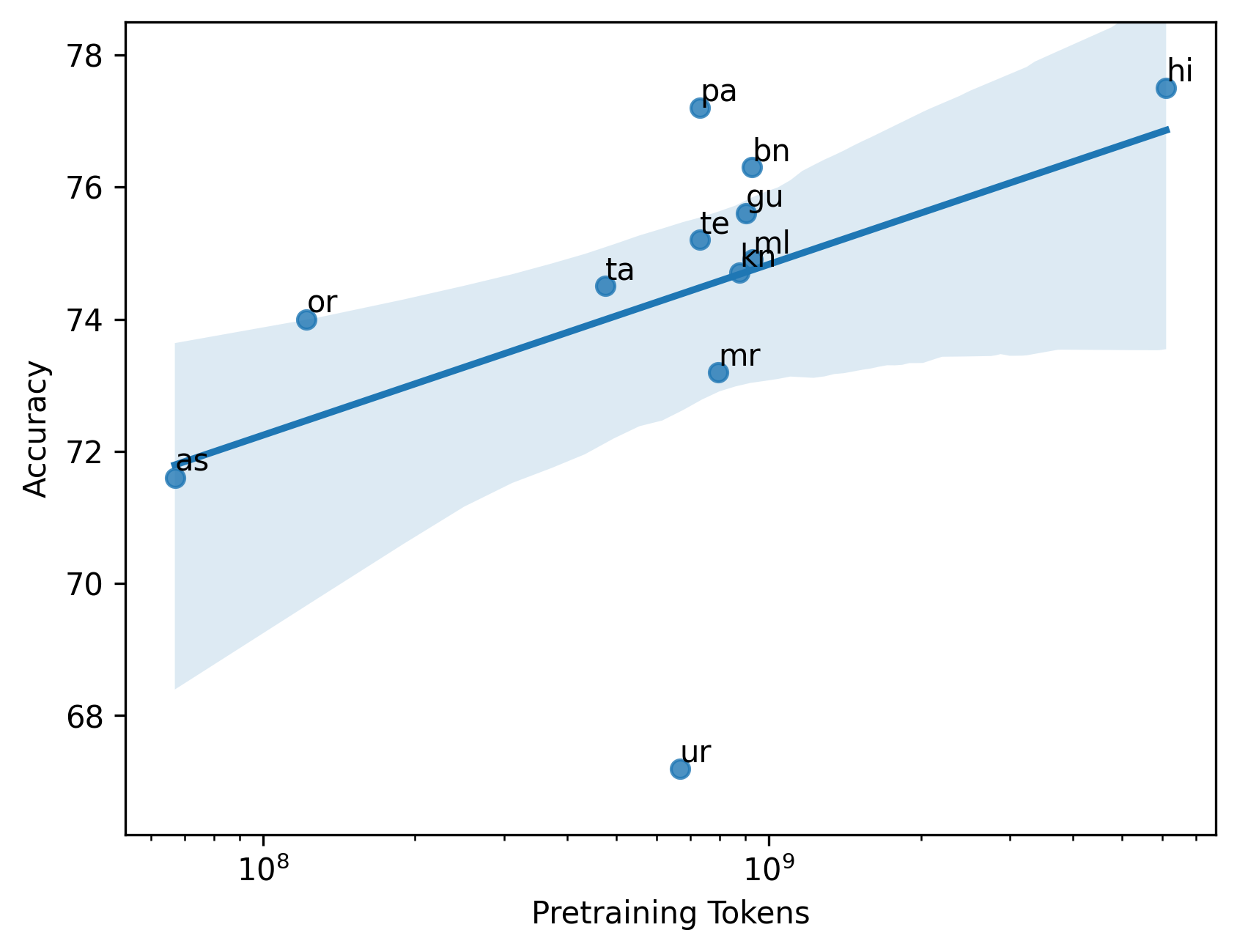}
         \caption{IndicXNLI}
     \end{subfigure}
     \hfill
     \begin{subfigure}[b]{0.3\textwidth}
         \centering
         \includegraphics[width=\textwidth]{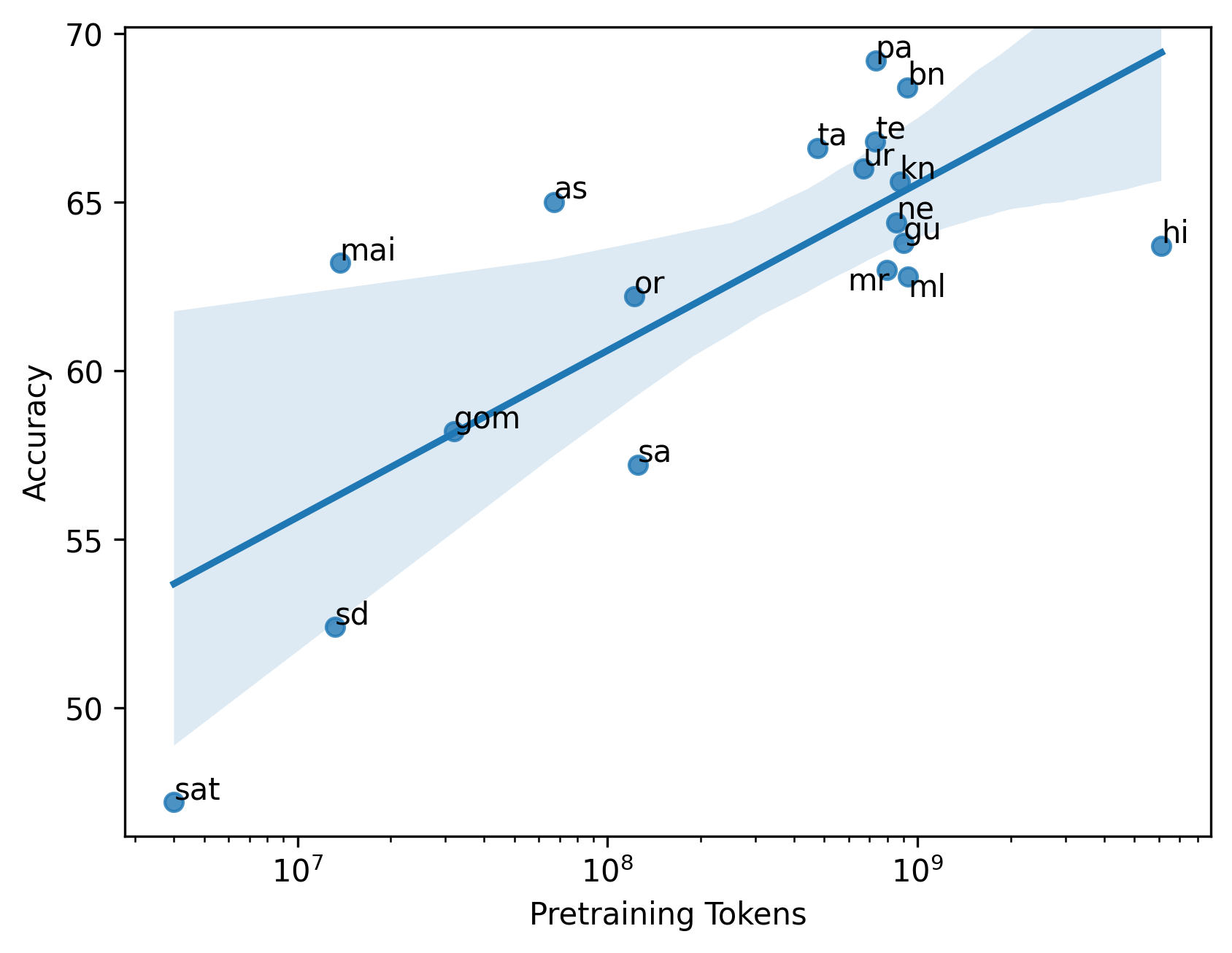}
         \caption{IndicCOPA}
     \end{subfigure}
     \hfill
     \begin{subfigure}[b]{0.3\textwidth}
         \centering
         \includegraphics[width=\textwidth]{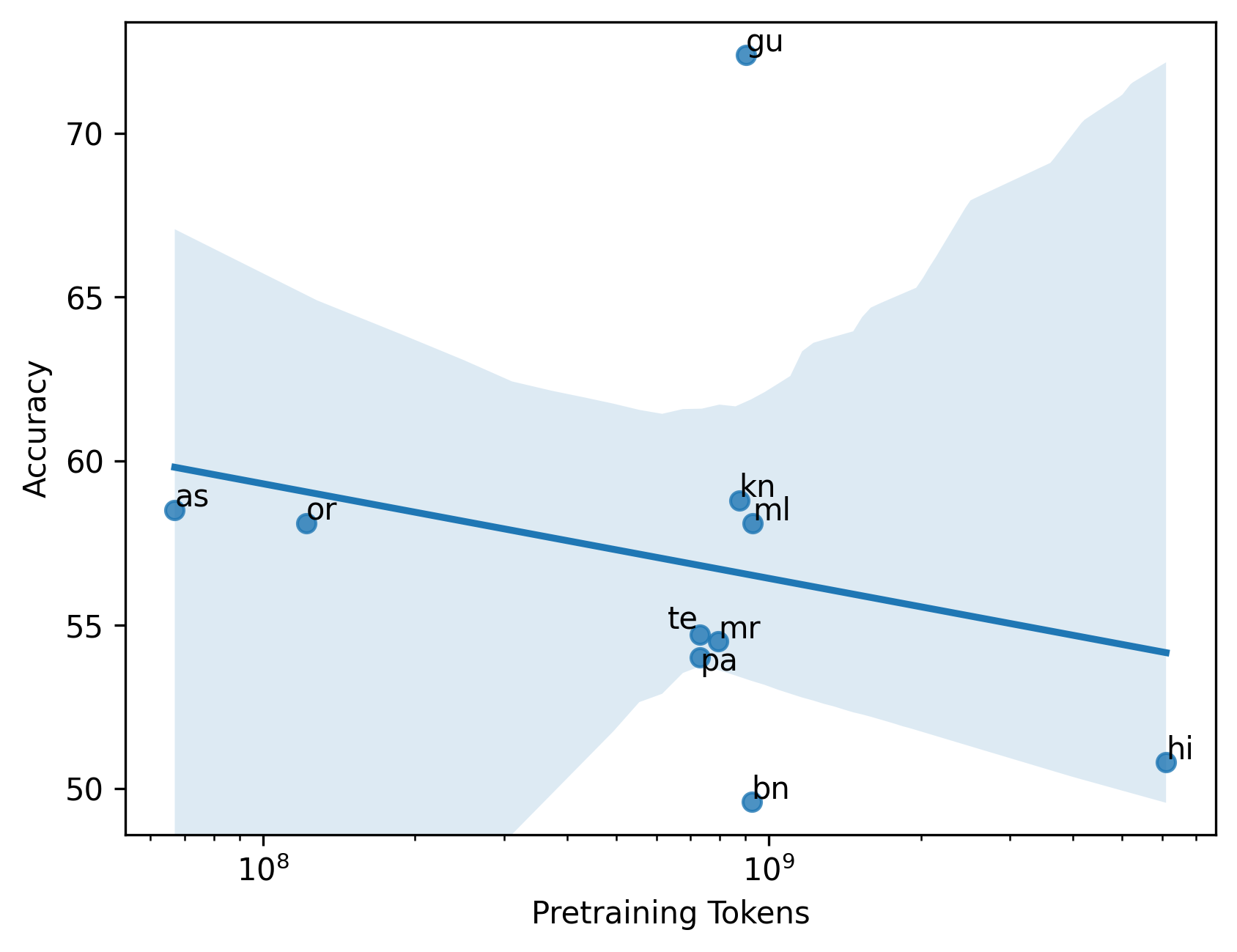}
         \caption{IndicXParaphrase}
         
     \end{subfigure}
     \hfill
     \begin{subfigure}[b]{0.3\textwidth}
         \centering
         \includegraphics[width=\textwidth]{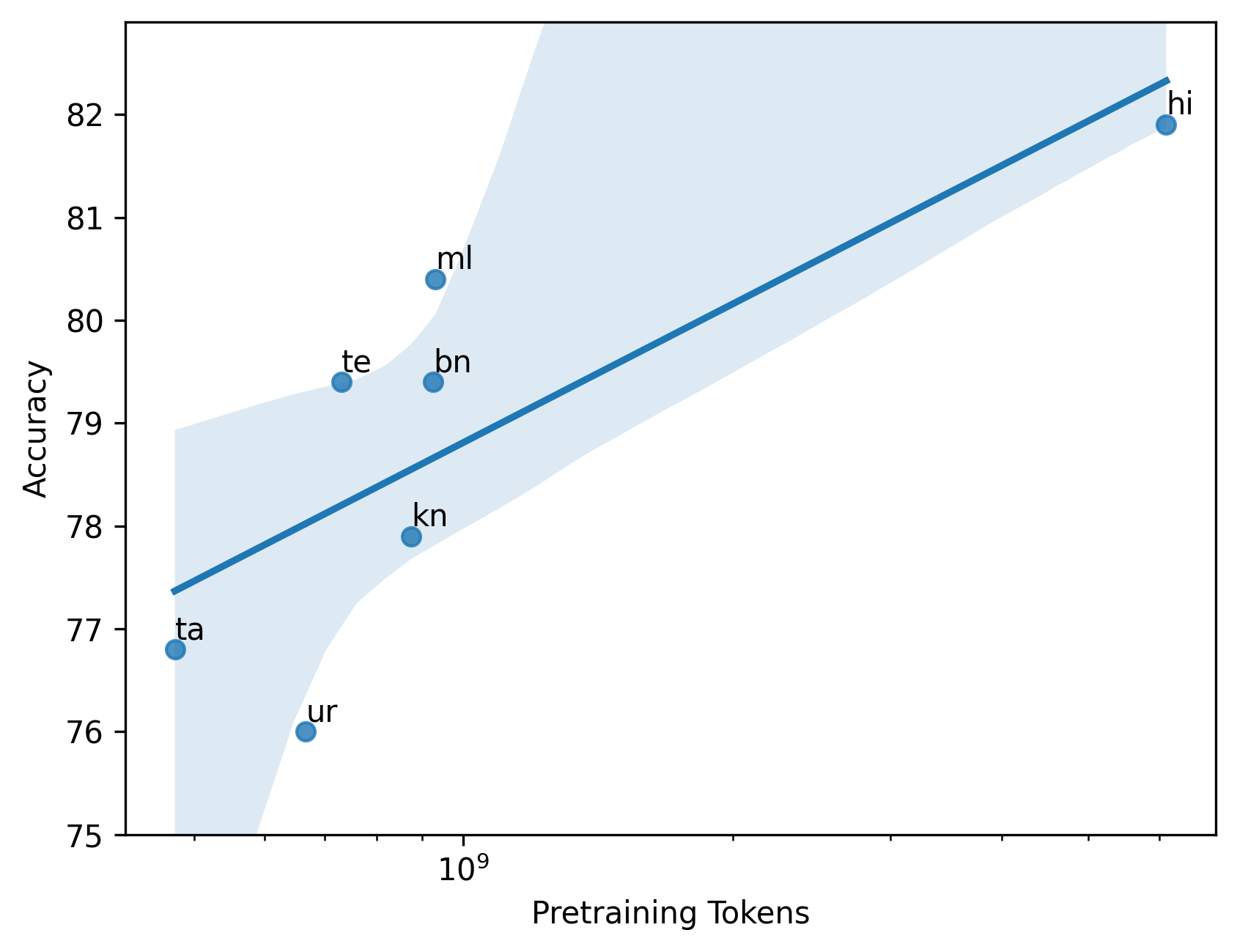}
         \caption{M-Intent}
     \end{subfigure}
     \hfill
     \begin{subfigure}[b]{0.3\textwidth}
         \centering
         \includegraphics[width=\textwidth]{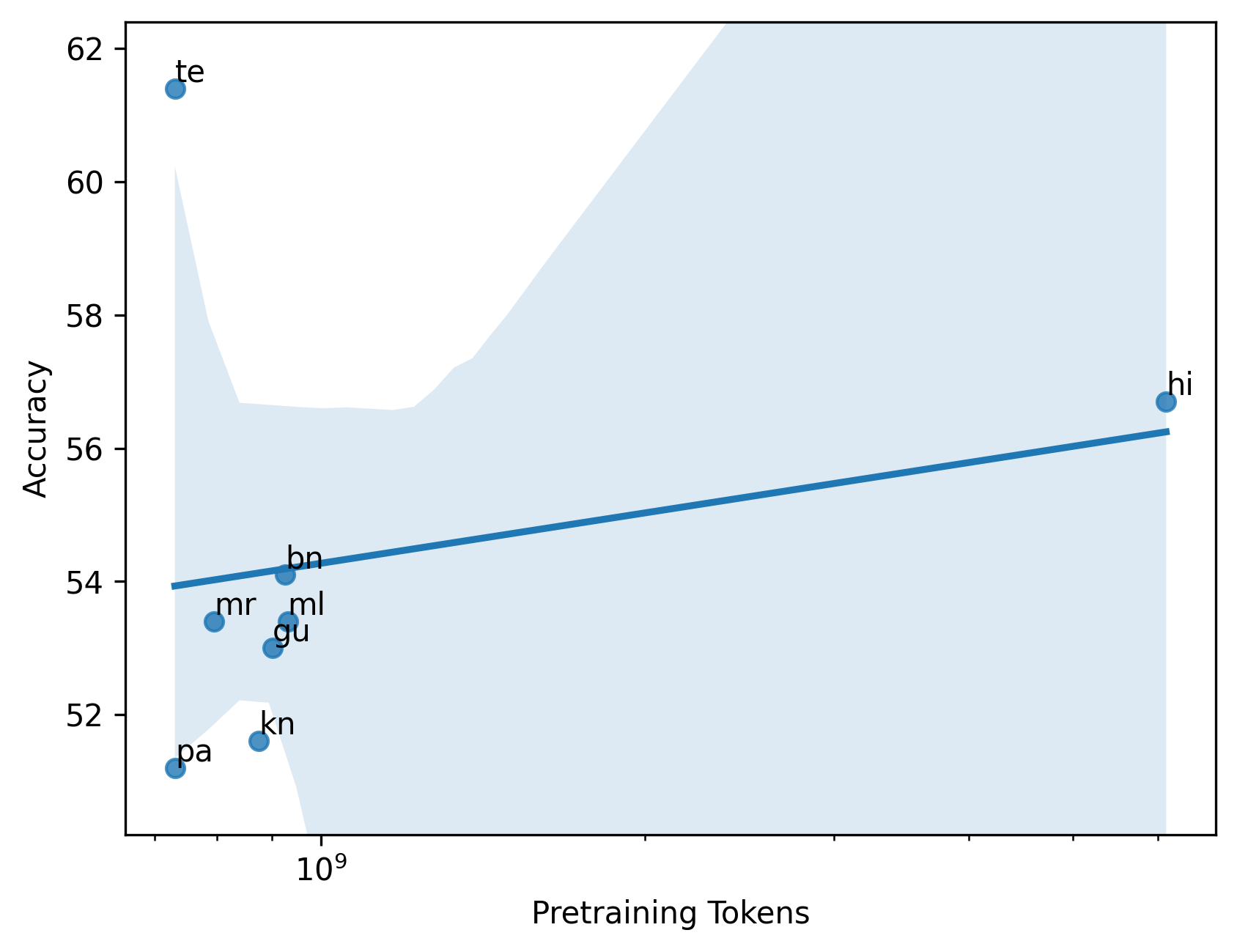}
         \caption{Naamapadam}
     \end{subfigure}
     \hfill
     \begin{subfigure}[b]{0.3\textwidth}
         \centering
         \includegraphics[width=\textwidth]{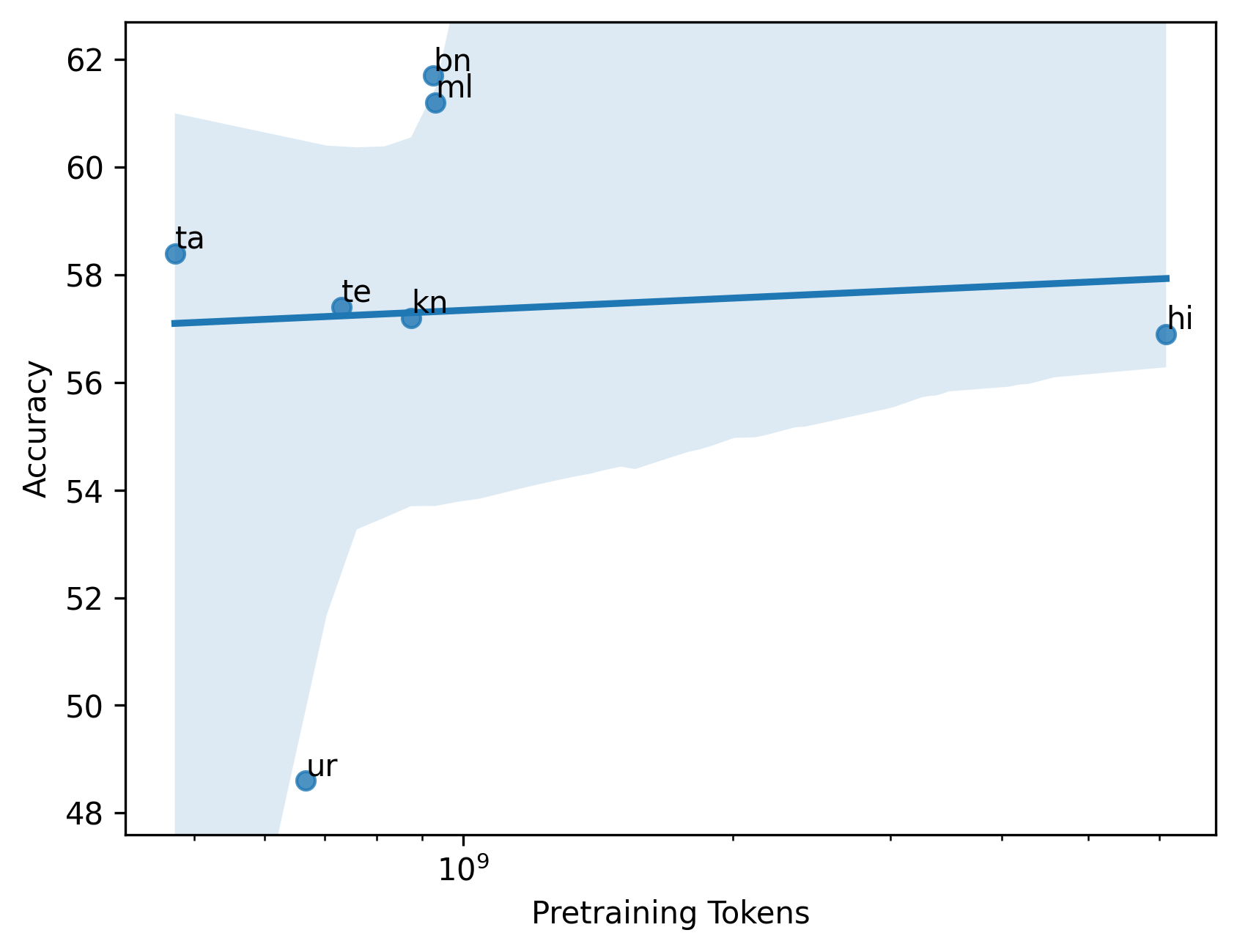}
         \caption{M-SlotFill}
     \end{subfigure}
     \hfill
     \begin{subfigure}[b]{0.3\textwidth}
         \centering
         \includegraphics[width=\textwidth]{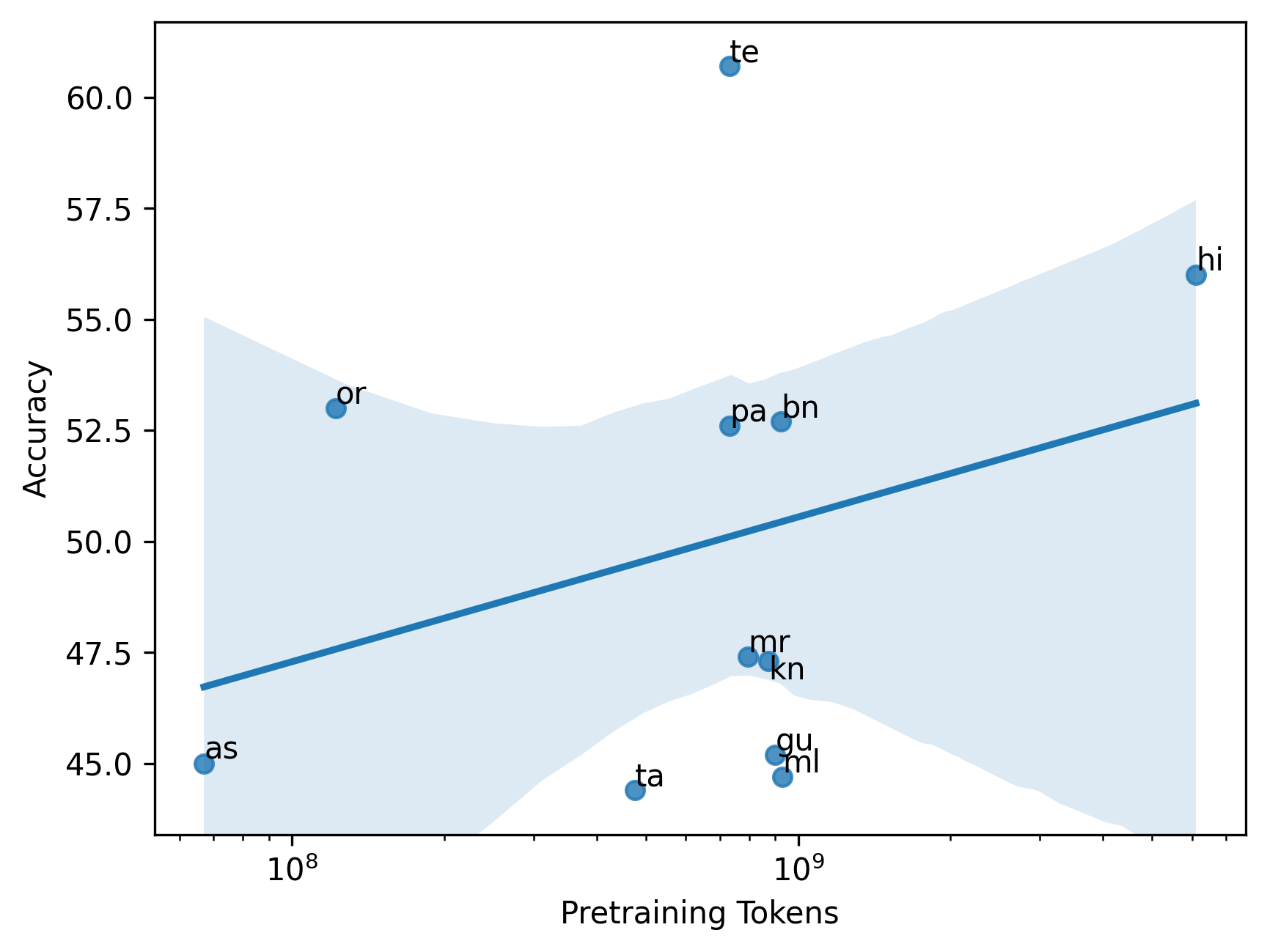}
         \caption{IndicQA}
     \end{subfigure}
     \hfill
     \begin{subfigure}[b]{0.3\textwidth}
         \centering
         \includegraphics[width=\textwidth]{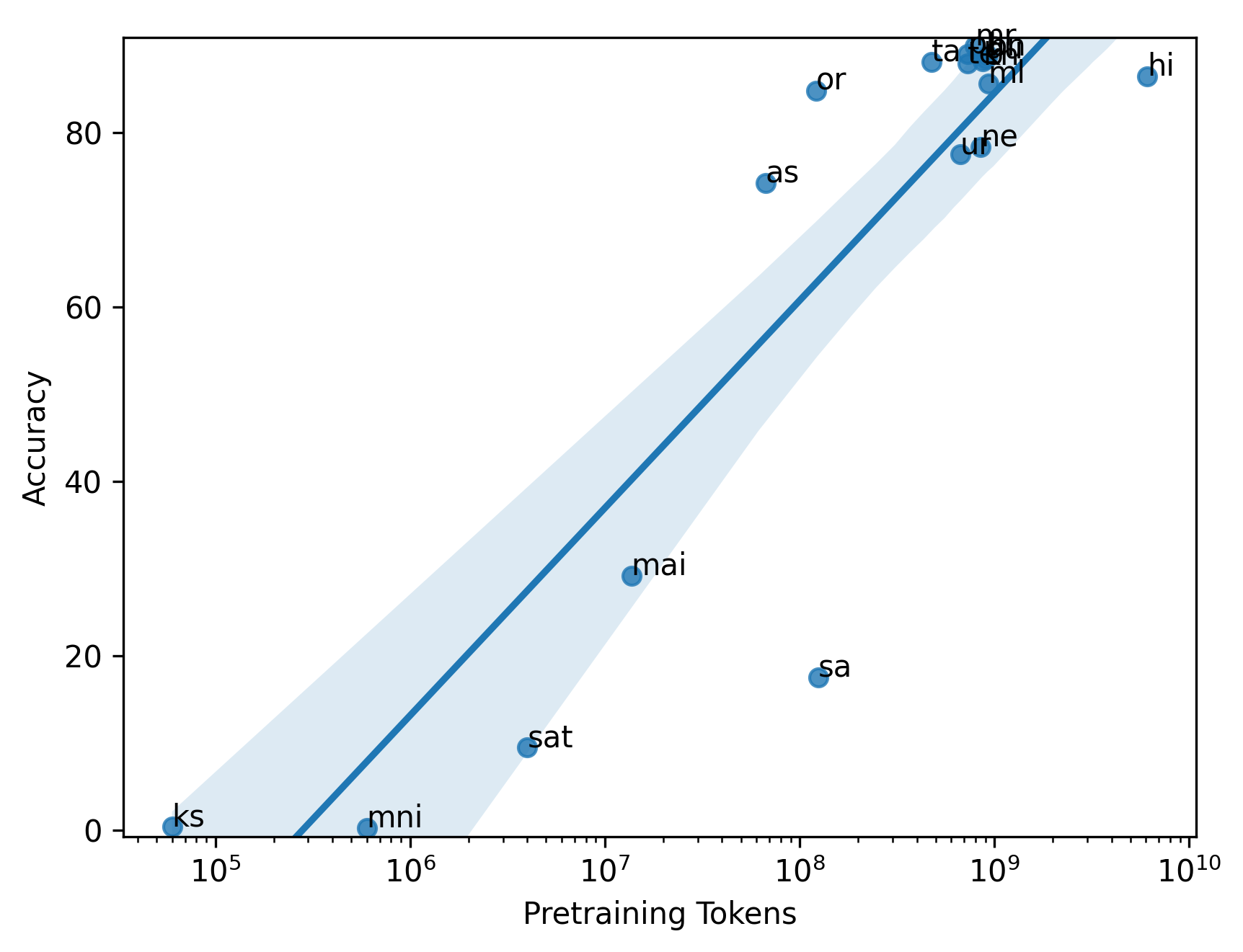}
         \caption{FLORES}
     \end{subfigure}
    %  \hfill
    %  \begin{subfigure}[b]{0.3\textwidth}
    %      \centering
    %      \includegraphics[width=\textwidth]{plots/10_avg_.png}
    %      \caption{Pivot Table \sd{need to fix} }
    %  \end{subfigure}
        \caption{Trends of pre-training data vs. downstream performance.}
        \label{fig:trends}
\end{figure*}

\section{Impact of pre-training data size} 
\label{app:pretraining_impact}
As expected, we can see from Fig.\ref{fig:trends} that as the size of pretraining data increases, there is an increase in downstream performance as well. This holds for all tasks across languages, except for IndicXParaphrase. It just holds for Naamapadam (NER) albeit with a high variance. As mentioned in Section \ref{sec:results}, we hypothesize that this could be due to the model's inability to learn good representations for noun phrases which play a major role in resolving named entities and paraphrase detection.

\begin{table}[]
\centering
\small
% \resizebox{\columnwidth}{!}{%
\begin{tabular}{llccc}
\toprule
 &  & \multicolumn{3}{c}{Best EN} \\
 \cmidrule(lr){3-5}
Task & Model & lr & wd & B* \\
 \midrule
\multirow{6}{*}{\shortstack{Indic-\\COPA}} & mBERT & 1e-05 & 0 & 3 \\
 & XLMR & 1e-05 & 0.01 & 4 \\
 & MuRIL & 3e-05 & 0 & 3 \\
 & IndicBERT & 3e-05 & 0.01 & 5 \\
 & $\quad$ +Samanantar & 3e-05 & 0 & 3 \\
 & $\quad\quad$ +Back-Trans & 3e-05 & 0 & 5 \\
 \midrule
\multirow{6}{*}{\shortstack{Indic-\\Paraphrase}} & mBERT & 3e-05 & 0.01 & 5 \\
 & XLMR & 1e-05 & 0.01 & 5 \\
 & MuRIL & 3e-05 & 0 & 3 \\
 & IndicBERT & 3e-05 & 0 & 3 \\
 & $\quad$ +Samanantar & 3e-05 & 0.01 & 5 \\
 & $\quad\quad$ +Back-Trans & 1e-05 & 0.01 & 5 \\
 \midrule
\end{tabular}
% }
\caption{Best hyperparameter configurations for IndicCOPA and IndicXParaphrase; lr, wd, and B* stand for learning rate, weight decay, and best epoch respectively.}
\label{tab:hparams-copa-para}
\end{table}

\begin{table*}[]
\centering
\small
\begin{tabular}{clccccccccc}
\toprule
 &  & \multicolumn{3}{c}{Best EN} & \multicolumn{3}{c}{Best IN} & \multicolumn{3}{c}{Best FAM} \\
 \cmidrule(lr){3-5} \cmidrule(lr){6-8} \cmidrule(lr){9-11}
Task & Model & lr & wd & B* & lr & wd & B* & lr & wd & B* \\
 \midrule
\multirow{6}{*}{\shortstack{Indic-\\Sentiment}} & mBERT & 3e-05 & 0.01 & 2 & 5e-06 & 0 & 2 & 5e-06 & 0 & 2 \\
 & XLMR & 5e-06 & 0.01 & 5 & 1e-05 & 0 & 2 & 5e-06 & 0 & 1 \\
 & MuRIL & 3e-05 & 0 & 4 & 5e-06 & 0.01 & 2 & 5e-06 & 0.01 & 1 \\
 & IndicBERT & 1e-05 & 0.01 & 3 & 1e-05 & 0.01 & 2 & 1e-05 & 0.01 & 2 \\
 & $\quad$ +Samanantar & 3e-05 & 0 & 3 & 3e-05 & 0 & 2 & 3e-05 & 0 & 2 \\
 & $\quad\quad$ +Back-Trans & 1e-05 & 0.01 & 5 & 1e-05 & 0 & 2 & 5e-06 & 0 & 2 \\
 \midrule
\multirow{6}{*}{\shortstack{Indic-\\XNLI}} & mBERT & 3e-05 & 0 & 3 & 5e-06 & 0 & 2 & 5e-06 & 0 & 2 \\
 & XLMR & 1e-05 & 0.01 & 5 & 1e-05 & 0 & 2 & 5e-06 & 0 & 4 \\
 & MuRIL & 3e-05 & 0.01 & 5 & 3e-05 & 0 & 2 & 3e-05 & 0 & 2 \\
 & IndicBERT & 3e-05 & 0.01 & 4 & 3e-05 & 0.01 & 4 & 1e-05 & 0.01 & 4 \\
 & $\quad$ +Samanantar & 3e-05 & 0.01 & 3 & 3e-05 & 0.01 & 3 & 3e-05 & 0.01 & 3 \\
 & $\quad\quad$ +Back-Trans & 1e-05 & 0.01 & 5 & 3e-05 & 0.01 & 2 & 3e-05 & 0.01 & 2 \\
 \midrule
\multirow{6}{*}{\shortstack{Naama-\\padam}} & mBERT & 3e-05 & 0 & 9 & 1e-05 & 0 & 7 & 1e-05 & 0 & 10 \\
 & XLMR & 3e-05 & 0.01 & 9 & 1e-5 & 0 & 9 & 1e-05 & 0 & 9 \\
 & MuRIL & 3e-05 & 0.01 & 10 & 1e-05 & 0 & 10 & 3e-05 & 0.01 & 6 \\
 & IndicBERT & 3e-05 & 0 & 10 & 3e-05 & 0 & 8 & 3e-05 & 0 & 8 \\
 & $\quad$ +Samanantar & 3e-05 & 0.01 & 6 & 3e-05 & 0 & 7 & 3e-05 & 0 & 7 \\
 & $\quad\quad$ +Back-Trans & 3e-05 & 0 & 10 & 3e-05 & 0.01 & 10 & 3e-05 & 0.01 & 10 \\
 \midrule
\multirow{6}{*}{\shortstack{Indic-\\QA}} & mBERT & 1e-05 & 0.01 & 4 & - & - & - & 1e-05 & 0.01 & 1 \\
 & XLMR & 2e-05 & 0.01 & 5 & - & - & - & 3e-05 & 0.01 & 5 \\
 & MuRIL & 3e-05 & 0.01 & 3 & - & - & - & 3e-05 & 0.01 & 5 \\
 & IndicBERT & 3e-05 & 0.01 & 4 & - & - & - & 3e-05 & 0.01 & 3 \\
 & $\quad$ +Samanantar & 3e-05 & 0 & 3 & - & - & - & 3e-05 & 0 & 2 \\
 & $\quad\quad$ +Back-Trans & 3e-05 & 0 & 5 & - & - & - & 3e-05 & 0 & 5 \\
 \bottomrule
\end{tabular}
\caption{Best hyperparameter configurations for datasets for which validation sets are available in  English, in-language, and in-language-family; lr, wd, and B* stand for learning rate, weight decay, and best epoch respectively.}
\label{tab:in-lang-hparams}
\end{table*}

\section{Fine-tuning Hyperparamters}
\label{app:ft-hparams}
We perform a grid search over learning rates [1e-5, 3e-5, 5e-6] and weight decay [0, 0.01] to choose the best model across tasks and languages. We report the best hyperparameters for English, in-language, and in-family validation sets. Table \ref{tab:hparams-copa-para} shows the best configuration for IndicCOPA and IndicXParaphrase for which only English validation sets are available. Table \ref{tab:in-lang-hparams} shows the best configurations for all other tasks for which both in-language and in-family validation sets are available.

For intent classification and slot-filling tasks, we use the same hyperparameter setting since they come from the same underlying data. We use a learning rate of 1e-5, weight decay of 0.1, and batch size of 256. For all the best models, unless otherwise mentioned we use a batch size of 32, and train with an initial warmup of 10\%. All the models are fine-tuned with half-precision on NVIDIA A100 GPUs.

\section{Language-wise Results}
\label{app:lang-specific}
Tables \ref{app-tab:sentiment}, \ref{app-tab:xnli}, \ref{app-tab:xcopa}, \ref{app-tab:xpara}, \ref{app-tab:m-intent}, \ref{app-tab:naamapadam}, \ref{app-tab:m-slot}, \ref{app-tab:qa}, \ref{app-tab:flores} show the language-wise results for IndicSentiment, IndicXNLI, IndicCOPA, IndicXParaphrase, MASSIVE Intent Classification, Naamapadam, MASSIVE Slot-filling, IndicQA, and FLORES sentence retrieval tasks respectively.

\begin{table*}[]
\centering
\begin{tabular}{llllcc}
\toprule
Code & Language & Script & Family & Class & Inclusivity \\
\midrule
as & Assamese & Bengali  & Indo-European & 2 & \cmark \\
brx & Bodo & Devanagari & Sino-Tibetan & 1 & \xmark \\
bn & Bengali & Bengali & Indo-European & 5 & \cmark \\
doi & Dogri & Devanagari & Indo-European & 1 & \xmark \\
en & English & Latin & Germanic & 5 & \cmark \\
gom & Konkani & Devanagari & Indo-European & 1 & \xmark \\
gu & Gujarati & Gujarati & Indo-European & 4 & \cmark \\
hi & Hindi & Devanagari & Indo-European & 5 & \cmark \\
kha & Khasi & Latin & Austroasiatic & 1 & \xmark \\
kn & Kannada & Kannada & Dravidian & 4 & \cmark \\
ks & Kashmiri & Arabic & Indo-European & 1 & \xmark \\
mai & Maithili & Devanagari & Indo-European & 1 & \xmark \\
ml & Malayalam & Malayalam & Dravidian & 4 & \cmark \\
mni & Manipuri & Meithi & Sino-Tibetan & 1 & \xmark \\
mr & Marathi & Devanagari  & Indo-European & 4 & \cmark \\
ne & Nepali & Devanagari  & Indo-European & 2 & \xmark \\
or & Odia & Odia & Indo-European & 3 & \cmark \\
pa & Punjabi & Gurumukhi & Indo-European & 3 & \cmark \\
sa & Sanskrit & Devanagari & Indo-European & 2 & \xmark \\
sat & Santali & Ol Chiki & Austroasiatic & 1 & \xmark \\
sd & Sindhi & Arabic & Indo-European & 1 & \xmark \\
ta & Tamil & Tamil & Dravidian & 4 & \cmark \\
te & Telugu & Telugu & Dravidian & 4 & \cmark \\
ur & Urdu & Arabic & Indo-European & 5 & \cmark \\
\bottomrule
\end{tabular}
\caption{Information about the languages present in \corpus/: their language family, class in the taxonomy introduced by \citet{joshi-etal-2020-state}, and inclusivity in other pre-trained models.}
\label{tab:lang-classes}
\end{table*}

\section{Language Classes}
Table \ref{tab:lang-classes} contains more information about each language in IndicCorp. We want to emphasize the diversity present in the corpus, and the differences in the size of resources available across languages through the classes to which they are assigned by \citet{joshi-etal-2020-state}.

% \usepackage{multirow}

\begingroup
\renewcommand{\arraystretch}{1} % Default value: 1
\setlength{\tabcolsep}{4pt} % Default value: 6pt
\begin{table*}[]
\small
\centering
\begin{tabular}{lcccccccccccccc}
\toprule
 & as & bd & bn & gu & hi & kn & ml & mr & or & pa & ta & te & ur & Avg. \\
\midrule
mBERT & 57.1 & 49.5 & 68.6 & 66.9 & 73.6 & 68.9 & 68.0 & 69.2 & 49.2 & 75.2 & 71.1 & 66.6 & 73.7 & 66.0 \\
XLMR & 80.2 & 51.6 & 88.7 & 85.1 & 89.3 & 86.8 & 86.7 & 89.3 & 84.3 & 86.4 & 87.8 & 88.4 & 87.0 & 84.0 \\
MuRIL & 87.8 & 48.8 & 90.8 & 85.9 & 90.6 & 87.5 & 86.0 & 90.4 & 87.0 & 88.0 & 88.9 & 87.4 & 89.9 & 85.3 \\
v1-data	& 90.9 & 60.2 & 92.7 & 91.9 & 92.2 & 90.6 & 90.1 & 91.9 & 88.2 & 90.6 & 90.6 & 91.6 & 52.9 & 85.7 \\
\midrule
IndicBERT & 91.4 & 80.4 & 91.8 & 90.5 & 91.4 & 90.1 & 90.3 & 91.7 & 90.7 & 91.6 & 92.3 & 91.6 & 89.0 & 90.2 \\
$\quad$ +Samanantar & \textbf{93.1} & \textbf{87.8} & \textbf{93.0} & \textbf{93.3} & \textbf{93.3} & \textbf{92.8} & \textbf{93.2} & \textbf{93.8} & \textbf{93.1} & \textbf{93.3} & \textbf{93.6} & \textbf{93.7} & \textbf{92.0} & \textbf{92.8} \\
$\quad\quad$ +Back-Trans. & 91.0 & 82.7 & 92.5 & 92.5 & 92.8 & 91.0 & 89.8 & 92.9 & 91.2 & 92.7 & 92.6 & 90.1 & 91.8 & 91.0 \\
\midrule
IndicBERT-SS & 92.0 & 89.7 & 91.2 & 91.8 & 92.2 & 90.6 & 91.5 & 91.6 & 91.9 & 92.4 & 91.4 & 91.3 & 91.4 & 91.5 \\
\bottomrule
\end{tabular}
\caption{Results on IndicSentiment task. Metric: accuracy.}
\label{app-tab:sentiment}
\end{table*}
\endgroup

\begin{table*}[]
\small
\centering
\begin{tabular}{lccccccccccccc}
\toprule
 & as & bn & gu & hi & kn & ml & mr & or & pa & ta & te & ur & Avg. \\
 \midrule
mBERT & 46.4 & 59.5 & 56.1 & 63.9 & 58.6 & 55.0 & 54.3 & 34.0 & 58.8 & 57.3 & 56.0 & 56.7 & 54.7 \\
XLMR & 63.5 & 70.7 & 70.5 & 75.2 & 71.5 & 71.3 & 69.0 & 68.5 & 70.1 & 70.7 & 69.6 & 65.3 & 69.7 \\
MuRIL & 70.1 & 74.5 & 73.1 & 76.3 & 74.0 & 71.8 & 70.6 & 70.8 & 74.8 & 72.9 & 72.7 & \textbf{67.6} & 72.4 \\
v1-data & 67.0 & 70.4 & 70.4 & 72.3 & 69.6 & 67.5 & 68.2 & 69.0 & 71.1 & 68.5 & 68.6 & 34.0 & 66.4 \\
\midrule
IndicBERT & 70.4 & 74.3 & 74.4 & 76.0 & 73.8 & 73.9 & 72.1 & 72.6 & 76.2 & 73.9 & 72.9 & 65.7 & 73.0 \\
$\quad$ +Samanantar & \textbf{71.6} & \textbf{76.3} & \textbf{75.6} & \textbf{77.5} & \textbf{74.7} & \textbf{74.9} & \textbf{73.2} & \textbf{74.0} & \textbf{77.2} & \textbf{74.5} & \textbf{75.2} & 67.2 & \textbf{74.3} \\
$\quad\quad$ +Back-Trans & 66.6 & 69.9 & 71.5 & 72.0 & 71.4 & 70.7 & 68.2 & 69.2 & 72.3 & 70.4 & 70.6 & 63.6 & 69.7 \\
\midrule
IndicBERT-SS & 70.9 & 76.0 & 76.0 & 77.8 & 75.3 & 73.5 & 72.3 & 74.2 & 76.1 & 73.7 & 74.3 & 66.9 & 73.9 \\
\bottomrule
\end{tabular}
\caption{Results on IndicXNLI task. Metric: accuracy.}
\label{app-tab:xnli}
\end{table*}

\begin{table*}[]
\centering
\small
\begin{tabular}{lcccccccccc}
\toprule
 & as & bn & gom & gu & hi & kn & mai & ml & mr & ne \\
\midrule
mBERT & 53.6 & 52.0 & 50.2 & 51.6 & 49.2 & 49.0 & 54.5 & 48.4 & 52.1 & 48.2 \\
XLMR & 58.0 & 62.6 & 56.4 & 60.7 & 59.9 & 60.8 & 56.6 & 59.4 & 58.4 & 58.8 \\
MuRIL & 60.2 & 63.0 & 52.0 & 60.7 & 57.7 & 61.6 & 57.2 & 58.2 & 56.3 & 57.0 \\
v1-data & 54.8 & 52.0 & 47.8 & 53.6 & 50.8 & 50.8 & 47.6 & 54.2 & 53.5 & 53.0 \\
\midrule
IndicBERT & 61.2 & \textbf{68.8} & \textbf{58.2} & 63.2 & 62.4 & \textbf{65.8} & 61.2 & 62.6 & \textbf{63.7} & 63.0 \\
$\quad$ +Samanantar & \textbf{65.0} & 68.4 & \textbf{58.2} & \textbf{63.8} & \textbf{63.7} & 65.6 & \textbf{63.2} & \textbf{62.8} & 63.0 & \textbf{64.4} \\
$\quad\quad$ +Back-Trans & 53.0 & 54.0 & 51.8 & 56.2 & 54.6 & 62.0 & 53.8 & 55.0 & 53.7 & 50.8 \\
\midrule
IndicBERT-SS & 65.0 & 69.0 & 63.4 & 64.5 & 63.0 & 67.6 & 61.8 & 64.0 & 64.1 & 59.6 \\
\midrule
 &  & or & pa & sa & sat & sd & ta & te & ur & Avg. \\
\midrule
mBERT &  & 48.8 & 51.8 & 47.2 & 52.0 & 50.6 & 51.8 & 51.8 & 56.2 & 51.7 \\
XLMR &  & 59.4 & 58.8 & 54.6 & 53.8 & 64.0 & 64.8 & 61.2 & 64.8 & 60.1 \\
MuRIL &  & 61.0 & 62.0 & 56.4 & 49.8 & 58.0 & 62.6 & 59.8 & 60.0 & 58.9 \\
v1-data & & 53.8 & 55.0 & 47.0 & 50.6 & 53.0 & 54.8 & 50.8 & 55.0 & 52.4 \\
\midrule
IndicBERT &  & \textbf{62.8} & 67.0 & \textbf{57.6} & \textbf{48.2} & \textbf{59.2} & \textbf{67.2} & 65.4 & 64.8 & 62.7 \\
$\quad$ +Samanantar & & 62.2 & \textbf{69.2} & 57.2 & 47.2 & 52.4 & 66.6 & \textbf{66.8} & \textbf{66.0} & \textbf{63.0} \\
$\quad\quad$ +Back-Trans & & 52.0 & 56.0 & 51.8 & 48.0 & 51.0 & 55.8 & 55.2 & 51.4 & 53.8 \\
\midrule
IndicBERT-SS & & 66.2 & 64.6 & 57.4 & 50.0 & 63.4 & 70.0 & 66.2 & 66.8 & 64.2 \\
\bottomrule
\end{tabular}
\caption{Results on IndicCOPA task. Metric: accuracy.}
\label{app-tab:xcopa}
\end{table*}

\begin{table*}[]
\centering
\small
\begin{tabular}{lccccccccccc}
\toprule
 & as & bn & gu & hi & kn & ml & mr & or & pa & te & Avg. \\
\midrule
mBERT & 48.3 & 50.5 & 78.1 & 51.3 & 49.5 & 53.4 & 58.9 & 50.0 & 55.2 & 56.7 & 55.2 \\
XLMR & 53.0 & 50.1 & 80.3 & 50.4 & 53.5 & 55.7 & 54.5 & 55.9 & 57.4 & 56.3 & 56.7 \\
MuRIL & \textbf{60.0} & \textbf{51.5} & \textbf{86.1} & \textbf{52.7} & \textbf{60.7} & \textbf{59.8} & \textbf{59.4} & \textbf{59.7} & \textbf{59.4} & \textbf{58.7} & \textbf{60.8} \\
v1-data & 49.5 & 49.5 & 52.6 & 49.2 & 48.0 & 49.1 & 47.9 & 49.6 & 51.2 & 49.5 & 49.6 \\
\midrule
IndicBERT & 57.1 & 50.1 & 74.9 & 50.3 & 57.9 & 56.8 & 54.3 & 57.2 & 55.0 & 55.2 & 56.9 \\
$\quad$ +Samanantar & 58.5 & 49.6 & 72.4 & 50.8 & 58.8 & 58.1 & 54.5 & 58.1 & 54.0 & 54.7 & 57.0 \\
$\quad\quad$ +Back-Trans & 50.6 & 54.2 & 50.1 & 50.7 & 49.3 & 50.3 & 50.3 & 50.0 & 51.1 & 50.2 & 50.7 \\
\midrule
IndicBERT-SS & 56.3 & 49.5 & 71.2 & 50.7 & 56.2 & 55.2 & 56.8 & 56.1 & 55.5 & 55.9 & 56.4 \\
\bottomrule
\end{tabular}
\caption{Results on IndicXParaphrase task. Metric: accuracy.}
\label{app-tab:xpara}
\end{table*}

\begin{table*}[]
\centering
\small
\begin{tabular}{lcccccccc}
\toprule
 & bn & hi & kn & ml & ta & te & ur & Avg. \\
\midrule
mBERT & 16.9 & 20.6 & 10.8 & 7.0 & 11.0 & 11.3 & 15.1 & 13.2 \\
XLMR & 63.7 & 74.9 & 61.7 & 69.5 & 65.7 & 66.6 & 63.8 & 66.6 \\
MuRIL & 77.0 & 82.4 & 77.5 & 77.4 & 75.9 & 74.7 & 75.7 & 77.2 \\
v1-data & 31.3 & 32.9 & 30.0 & 29.7 & 25.5 & 30.5 & 1.1 & 25.8 \\
\midrule
IndicBERT & \textbf{79.5} & \textbf{82.7} & \textbf{78.2} & \textbf{80.4} & 76.1 & 77.9 & \textbf{76.9} & \textbf{78.8} \\
$\quad$ +Samanantar & 79.4 & 81.9 & 77.9 & 80.4 & \textbf{76.8} & \textbf{79.4} & 76.0 & \textbf{78.8} \\
$\quad\quad$ +Back-Trans & 79.1 & 81.0 & 77.2 & 79.5 & 75.6 & 76.7 & 73.1 & 77.4 \\
\midrule
IndicBERT-SS & 80.6 & 83.4 & 79.3 & 81.6 & 78.4 & 81.5 & 80.5 & 80.7 \\
\bottomrule
\end{tabular}
\caption{Results on MASSIVE Intent Classification task. Metric: accuracy.}
\label{app-tab:m-intent}
\end{table*}

% \begin{table*}[]
% \centering
% \small
% \begin{tabular}{lcccccccccc}
% \toprule
%  & bn & gu & hi & kn & ml & mr & pa & ta & te & Avg. \\
%  \midrule
% mBERT & 47.7 & 45.7 & 55.6 & 53.4 & 54.1 & 53.9 & 45.0 & 42.8 & 58.2 & 50.7 \\
% XLMR & 41.8 & 47.5 & 57.0 & 41.7 & 49.6 & 48.3 & 40.7 & 39.1 & 52.4 & 46.4 \\
% MuRIL & \textbf{58.7} & \textbf{57.9} & \textbf{59.9} & \textbf{59.7} & \textbf{64.2} & \textbf{55.5} & \textbf{51.8} & \textbf{46.8} & \textbf{67.0} & \textbf{57.9} \\
% \midrule
% IndicBERT & 53.6 & 51.1 & 54.0 & 51.0 & 53.6 & 51.5 & 48.9 & 44.5 & 63.2 & 52.4 \\
% $\quad$ +Samanantar & 52.4 & 50.9 & 55.5 & 52.1 & 52.6 & 52.4 & 49.3 & 43.0 & 60.3 & 52.0 \\
% $\quad\quad$ +Back-Trans & 54.1 & 55.3 & 56.4 & 53.4 & 56.8 & 53.8 & 51.0 & 45.9 & 64.0 & 54.5 \\
% \bottomrule
% \end{tabular}
% \caption{\sd{Results on Naamapadam NER task. Metric: F1 score.}}
% \label{app-tab:naamapadam}
% \end{table*}

\begin{table*}[]
\small
\centering
\begin{tabular}{lcccccccccc}
\toprule
 & bn & gu & hi & kn & ml & mr & pa & ta & te & Avg. \\
\midrule
mBERT & 61.1 & 55.4 & 70.9 & 64.1 & 63.9 & 67.1 & 57.4 & 57.7 & 69.0 & 63.0 \\
XLMR & 69.3 & 70.2 & 79.0 & 72.2 & 74.1 & 71.5 & 67.3 & 64.3 & 77.9 & 71.7 \\
MuRIL & 72.5 & \textbf{75.1} & \textbf{79.5} & \textbf{76.2} & \textbf{75.3} & \textbf{73.3} & 71.1 & \textbf{64.5} & \textbf{81.1} & \textbf{74.3} \\
v1-data & 60.7 & 58.6 & 61.9 & 58.4 & 60.1 & 53.1 & 55.1 & 51.3 & 65.4 & 58.3 \\
\midrule
IndicBERT & \textbf{74.1} & 72.5 & 78.5 & 74.8 & 72.5 & 71.7 & \textbf{71.4} & 63.7 & 79.8 & 73.2 \\
$\quad$ +Samanantar & 72.5 & 73.8 & 76.7 & 73.3 & 72.2 & 71.6 & 69.3 & 64.0 & 78.1 & 72.4 \\
$\quad\quad$ +Back-Trans & 71.6 & 72.4 & 76.4 & 73.6 & 71.7 & 71.0 & 67.6 & 63.7 & 78.7 & 71.9 \\
\midrule
IndicBERT-SS & 69.1 & 64.0 & 75.5 & 64.5 & 66.5 & 65.1 & 64.2 & 57.6 & 72.7 & 66.6 \\
\bottomrule
\end{tabular}
\caption{Results on Naamapadam NER task. Metric: F1 score.}
\label{app-tab:naamapadam}
\end{table*}

\begin{table*}[]
\centering
\small
\begin{tabular}{lcccccccc}
\toprule
 & bn & hi & kn & ml & ta & te & ur & Avg. \\
 \midrule
mBERT & 7.3 & 10.2 & 5.8 & 3.5 & 5.6 & 4.0 & 7.3 & 6.2 \\
XLMR & 51.4 & 55.9 & 48.1 & 52.3 & 50.2 & 51.3 & 41.1 & 50.0 \\
MuRIL & 60.5 & 57.5 & 55.9 & 58.6 & \textbf{58.5} & 57.0 & \textbf{51.0} & 57.0 \\
v1-data & 41.1 & 42.8 & 42.2 & 38.6 & 34.4 & 40.6 & 0.8 & 34.4 \\
\midrule
IndicBERT & 61.6 & 55.4 & 55.9 & 60.4 & 56.8 & \textbf{58.3} & 48.5 & 56.7 \\
$\quad$ +Samanantar & \textbf{61.7} & \textbf{56.9} & \textbf{57.2} & \textbf{61.2} & 58.4 & 57.4 & 48.6 & 57.3 \\
$\quad\quad$ +Back-Trans & 58.6 & 52.7 & 55.8 & 59.0 & 55.4 & 54.1 & 46.7 & 54.6 \\
\midrule
IndicBERT-SS & 58.9 & 54.7 & 57.9 & 61.0 & 58.1 & 59.2 & 51.0 & 57.3 \\
\bottomrule
\end{tabular}
\caption{Results on MASSIVE Slot-filling task. Metric: F1 score.}
\label{app-tab:m-slot}
\end{table*}

\begin{table*}[]
\small
\centering
% \resizebox{\textwidth}{!}{%
\begin{tabular}{lcccccccccccc}
\toprule
 & as & bn & gu & hi & kn & ml & mr & or & pa & ta & te & Avg. \\
 \midrule
mBERT & 18.2 & 42.1 & 29.9 & 41.1 & 37.0 & 32.2 & 36.1 & 3.9 & 39.3 & 33.1 & 48.8 & 32.9 \\
XLMR & 34.3 & 47.1 & 39.4 & 52.0 & 42.0 & 40.3 & 43.9 & 43.4 & 49.1 & 43.8 & 57.5 & 44.8 \\
MuRIL & 43.2 & 52.1 & 43.2 & 54.2 & 44.8 & \textbf{43.9} & \textbf{48.0} & 47.5 & 46.2 & 45.0 & 56.9 & 47.7 \\
v1-data & 30.8 & 39.7 & 35.8 & 37.7 & 34.7 & 36.2 & 38.9 & 37.6 & 39.8 & 34.4 & 48.1 & 37.6 \\
\midrule
IndicBERT & 44.5 & 51.6 & 43.8 & 54.7 & 45.9 & 43.7 & 46.3 & 47.2 & 51.1 & 43.5 & 59.1 & 48.3 \\
$\quad$ +Samanantar & \textbf{45.3} & \textbf{52.7} & \textbf{44.3} & \textbf{55.6} & \textbf{46.3} & \textbf{43.9} & 47.1 & \textbf{48.1} & \textbf{52.3} & \textbf{45.4} & \textbf{59.7} & \textbf{49.2} \\
$\quad\quad$ +Back-Trans & 37.3 & 47.0 & 37.8 & 48.0 & 39.1 & 35.1 & 38.5 & 41.7 & 47.5 & 39.8 & 52.3 & 42.2 \\
\midrule
IndicBERT-SS & 44.8 & 53.9 & 45.2 & 55.6 & 46.1 & 47.8 & 48.9 & 49.9 & 52.6 & 44.0 & 57.7 & 49.7 \\
\bottomrule
\end{tabular}
% }
\caption{Results on IndicQA task. Metric: F1 score.}
\label{app-tab:qa}
\end{table*}

\begin{table*}[]
\centering
\small
\begin{tabular}{lcccccccccc}
\toprule
 & as & bn & gu & hi & kn & ks & mai & ml & mr & mni \\
 \midrule
mBERT & 9.4 & 47.2 & 32.4 & 62.6 & 46.1 & \textbf{11.9} & 32.4 & 33.6 & 47.7 & \textbf{2.5} \\
XLMR & 0.3 & 3.3 & 2.9 & 9.6 & 3.7 & 0.3 & 0.8 & 1.9 & 7.0 & 0.3 \\
MuRIL & 40.3 & 77.0 & 67.0 & 84.2 & 88.4 & 9.3 & 16.3 & 82.2 & 83.9 & 0.7 \\
v1-data & 77.7 & 85.6 & 89.6 & 89.8 & 84.5 & 0.6 & 23.4 & 80.2 & 87.9 & 1.9 \\
\midrule
IndicBERT & \textbf{86.0} & 91.0 & \textbf{92.4} & 90.5 & 89.1 & 0.9 & 38.1 & \textbf{89.2} & 92.5 & 0.3 \\
$\quad$ +Samanantar & 74.2 & 88.8 & 88.4 & 86.4 & 88.2 & 0.4 & 29.2 & 85.6 & 89.9 & 0.3 \\
$\quad\quad$ +Back-Trans & 79.2 & \textbf{91.1} & 90.5 & \textbf{94.3} & \textbf{89.8} & 1.8 & \textbf{41.9} & 88.1 & \textbf{94.0} & 0.5 \\
\midrule
IndicBERT-SS & 85.5 & 92.0 & 85.5 & 84.8 & 87.7 & 2.1 & 79.2 & 91.7 & 85.5 & 0.2 \\
\midrule
 &  & ne & or & pa & sa & sat & ta & te & ur & Avg. \\
\midrule
mBERT &  & 54.7 & 2.3 & 38.0 & 14.5 & 0.7 & 47.4 & 40.3 & 57.7 & 32.3 \\
XLMR &  & 8.9 & 2.8 & 0.7 & 1.5 & 0.0 & 5.0 & 4.5 & 2.2 & 3.1 \\
MuRIL &  & 59.1 & 37.1 & 71.9 & 36.4 & 0.5 & 79.4 & 43.5 & 65.1 & 52.3 \\
v1-data & & 16.0 & 82.9 & 88.3 & 9.5 & 0.7 & 83.9 & 84.7 & 0.2 \\
\midrule
IndicBERT & \textbf{} & \textbf{79.9} & \textbf{90.9} & \textbf{92.2} & 30.4 & \textbf{19.9} & 90.0 & 88.6 & \textbf{87.0} & \textbf{69.4} \\
$\quad$ +Samanantar &  & 78.3 & 84.8 & 89.0 & 17.5 & 9.5 & 88.1 & 87.9 & 77.5 & 64.7 \\
$\quad\quad$ +Back-Trans &  & 75.8 & 85.8 & 90.5 & \textbf{40.9} & 7.8 & \textbf{90.5} & \textbf{89.3} & 82.6 & 68.6 \\
\midrule
IndicBERT-SS & & 73.8 & 90.8 & 92.9 & 36.9 & 24.9 & 89.2 & 86.5 & 92.3 \\
\bottomrule
\end{tabular}
\caption{Results on FLORES sentence retrieval task. Metric: accuracy.}
\label{app-tab:flores}
\end{table*}

\begin{table*}[]
\small
\centering
\begin{tabular}{lccccccccccc}
\toprule
 & as & bd & bn & gom & gu & hi & kn & ks & ml & mai & mr \\
 \midrule
mBERT & 38.8 & 49.5 & 43.5 & 50.2 & 51.5 & 47.6 & 42.1 & 11.9 & 39.5 & 43.5 & 53.2 \\
XLMR & 48.2 & 51.6 & 53.3 & 56.4 & 55.2 & 58.2 & 52.2 & 0.3 & 54.1 & 28.7 & 52.9 \\
MuRIL & 60.3 & 48.8 & 67.2 & 52.0 & 67.7 & 68.4 & 67.8 & 9.3 & 66.9 & 36.8 & 66.3 \\
v1-data & 61.8 & 60.2 & 57.2 & 47.8 & 63.6 & 58.1 & 55.6 & 0.6 & 54.9 & 35.5 & 61.9 \\
\midrule
IndicBERT & 68.4 & 80.4 & 69.1 & 58.2 & 70.0 & 68.6 & 67.5 & 0.9 & 67.9 & 49.7 & 67.4 \\
$\quad$ +Samanantar & 68.0 & 87.8 & 69.2 & 58.2 & 69.8 & 69.1 & 68.2 & 0.4 & 68.1 & 46.2 & 67.7 \\
$\quad\quad$ +Back-Trans & 63.0 & 82.7 & 66.7 & 51.8 & 64.8 & 66.9 & 65.4 & 1.8 & 64.9 & 47.8 & 64.5 \\
\midrule
IndicBERT-SS & 69.1 & 89.7 & 69.4 & 63.4 & 68.8 & 68.2 & 67.8 & 2.1 & 68.6 & 70.5 & 66.9 \\
\midrule
 &  & mni & or & pa & sa & sat & sd & ta & te & ur & avg \\
 \midrule
mBERT &  & 2.5 & 31.4 & 51.9 & 30.9 & 26.3 & 50.6 & 40.0 & 43.7 & 44.4 & 39.6 \\
XLMR &  & 0.3 & 52.4 & 51.9 & 28.0 & 26.9 & 64.0 & 53.4 & 56.4 & 54.0 & 44.9 \\
MuRIL &  & 0.7 & 60.5 & 64.9 & 46.4 & 25.1 & 58.0 & 66.2 & 64.2 & 68.2 & 53.3 \\
v1-data &  & 1.9 & 63.5 & 63.4 & 28.2 & 25.6 & 53.0 & 53.6 & 58.0 & 24.0 & 46.4 \\
\midrule
IndicBERT &  & 0.3 & 70.2 & 68.9 & 44.0 & 34.0 & 59.2 & 68.0 & 70.2 & 72.0 & 57.8 \\
$\quad$ +Samanantar &  & 0.3 & 70.0 & 69.2 & 37.3 & 28.3 & 52.4 & 68.3 & 70.6 & 71.2 & 57.0 \\
$\quad\quad$ +Back-Trans &  & 0.5 & 65.0 & 65.9 & 46.4 & 27.9 & 51.0 & 65.8 & 66.9 & 68.2 & 54.9 \\
\midrule
IndicBERT-SS &  & 0.2 & 71.5 & 68.4 & 47.1 & 37.5 & 63.4 & 68.3 & 70.1 & 74.8 & 60.3 \\
\bottomrule
\end{tabular}
\caption{Results averaged across \textbf{tasks} using preferred metric from the \benchmark/ benchmark.}
\label{tab:lang-wise-results}
\end{table*}

% \begin{table}[h!]
% \centering
% \small
% % \resizebox{\columnwidth}{!}{
% \begin{tabular}{lcccc}
% \toprule
% % & en-in & en-(in-fam) & hi-in & hi-(in-fam) \\
% & \multicolumn{2}{c}{en} & \multicolumn{2}{c}{hi} \\
% \cmidrule(lr){2-3} \cmidrule(lr){4-5}
% & in-lg. & in-fam. & in-lg. & in-fam. \\
%  \midrule
% mBERT & 55.6 & 55.4 & 71.9 & 72.0 \\
% XLMR & 58.6 & 57.7 & 76.7 & 76.5 \\
% MuRIL & 64.7 & 64.7 & 78.3 & 77.8 \\
% \midrule
% IndicBERT & 58.6 & 58.6 & 78.3 & 78.3 \\
% $\quad$ +Samanantar & 57.5 & 57.5 & 78.2 & 77.4 \\
% $\quad\quad$ +Back-Trans. & 57.9 & 58.0 & 77.3 & 77.3 \\
% \bottomrule
% \end{tabular}
% % }
% \caption{Naamapadam ``transfer-language" experiment. The columns indicate the fine-tuning-development language combination. We restrict the size of the Hindi fine-tuning set to 20,000 examples to match the size of the English set. We remove English and Hindi testsets while computing the average to avoid skewing the averages.}
% \label{tab:hi-zero-shot}
% \end{table}

\end{document}